\documentclass[11pt]{article}
\usepackage[paper=letterpaper,margin=1in]{geometry}

\usepackage{natbib}

\usepackage[utf8]{inputenc} %
\usepackage[T1]{fontenc}    %
\usepackage{hyperref}       %
\usepackage{url}            %
\usepackage{booktabs}       %
\usepackage{amsfonts}       %
\usepackage{nicefrac}       %
\usepackage{microtype}      %

\usepackage{microtype}
\usepackage{subcaption}

\usepackage{amssymb,amsmath,amsfonts,amsthm,commath}
\usepackage{algorithm}
\usepackage{xspace}
\usepackage[capitalize]{cleveref}
\usepackage{breqn}
\usepackage{threeparttable}
\usepackage{mdframed}
\usepackage{algorithm}
\usepackage{algorithmic}
\usepackage{adjustbox}
\usepackage{balance}

\usepackage[pdftex,dvipsnames,table]{xcolor}  %
\newcommand{\revise}[1]{{#1}}
\DeclareMathOperator*{\argmax}{arg\,max}

\DeclareMathOperator*{\diam}{diam}

\newcommand{\tF}{\Tilde{F}}
\newcommand{\expect}{\mathbb{E}}
\newcommand{\constraint}{\mathcal{K}}
\newcommand{\domain}{\mathcal{D}}
\newcommand{\domainsh}{\domain_\delta}
\newcommand{\matroid}{\mathcal{I}}
\newcommand{\one}{\mathbf{1}}

\newcommand{\AlgCG}{\textsf{Continuous Greedy}\xspace}
\newcommand{\Alg}{\textsf{Black-box Continuous Greedy}\xspace}
\newcommand{\Algdis}{\textsf{Discrete Black-box Greedy}\xspace}

\newtheorem{theorem}{Theorem}
\newtheorem{lemma}{Lemma}

\newtheorem{remark}{Remark}

\begin{document}

\title{Black Box Submodular Maximization:\\ Discrete and Continuous Settings}

\author{ Lin Chen\thanks{Yale Institute for Network Science, Yale 
University. E-mail: \texttt{lin.chen@yale.edu}. First two authors contributed 
equally to this work.} 
\and
	 Mingrui Zhang\thanks{Department of Statistics and Data Science,
	 	Yale University. E-mail: \texttt{mingrui.zhang@yale.edu}}
	\and
	 Hamed Hassani\thanks{Department of Electrical and Systems Engineering,
	 	University of Pennsylvania. E-mail: \texttt{hassani@seas.upenn.edu}} 
	\and
	 Amin Karbasi\thanks{Yale Institute for Network Science, Yale University. 
	 E-mail: \texttt{amin.karbasi@yale.edu}} 
}

\date{}

\maketitle

\begin{abstract}
	In this paper, we consider the problem of black box continuous submodular 
	maximization 
	where we only have  access to the  function values and no information about 
	the 
	derivatives 
	is provided. For a 
	monotone and continuous DR-submodular function, and subject to a bounded 
	convex body 
	constraint, we  
	propose  \Alg, a derivative-free algorithm  that provably achieves the 
	tight 
	$[(1-1/e)OPT-\epsilon]$ approximation guarantee with $O(d/\epsilon^3)$ 
	function evaluations. 
	We then extend our result to the stochastic setting where function values 
	are subject to 
	stochastic zero-mean noise. It is through this stochastic generalization 
	that we revisit the discrete 
	submodular maximization problem and use the multi-linear extension as a 
	bridge between 
	discrete and continuous settings. Finally, we extensively 
	evaluate the 
	performance of our algorithm on  continuous and discrete submodular 
	objective 
	functions using both synthetic and real data. 
\end{abstract}

\section{Introduction}

Black-box optimization, also known as zeroth-order or derivative-free
optimization\footnote{\revise{We note that black-box 
optimization (BBO) and 
derivative-free 
optimization (DFO) are not identical terms. 
\citet{audet2017derivative} defined DFO as ``the mathematical study of 
optimization 
algorithms that do not use derivatives'' and BBO as ``the study of design 
and analysis of algorithms that assume the objective and/or constraint 
functions are given by blackboxes''. However, as the differences are nuanced in 
most scenarios, this paper uses them interchangeably.}}, has been 
extensively studied in the literature 
\citep{conn2009introduction,bergstra2011algorithms,rios2013derivative,shahriari2016taking}.
In this setting, we assume that the objective function is unknown 
and we can only obtain 
zeroth-order information such as  (stochastic) function evaluations. 

Fueled by a growing number of machine learning applications, black-box 
optimization  methods are usually 
considered in scenarios where 
gradients (\emph{i.e.}, 
first-order information) are 1) difficult or slow to compute, \emph{e.g.}, 
graphical 
model inference \citep{wainwright2008graphical}, structure predictions 
\citep{taskar2005learning,sokolov2016stochastic}, or 2) inaccessible, 
\emph{e.g.}, 
hyper-parameter 
turning for natural language processing or image 
classifications \cite{snoek2012practical,thornton2013auto}, black-box attacks 
for finding adversarial examples \cite{chen2017zoo,ilyas2018black}. Even though 
heuristics such as 
random or grid search, with undesirable dependencies on the dimension, are 
still used in some applications 
(\emph{e.g.}, 
parameter tuning for deep networks), there has been a growing number of 
rigorous methods to address the 
convergence rate of black-box optimization in convex and non-convex settings 
\citep{wang2017stochastic,balasubramanian2018zeroth,sahu2018towards}.

The focus of this paper is the \textit{constrained} continuous 
DR-submodular maximization %
over a bounded convex body. We aim to design an algorithm that uses only 
zeroth-order 
information while avoiding expensive  projection operations. Note that one way 
the 
optimization methods can deal with constraints is to apply the projection 
oracle once the 
proposed iterates land outside the feasibility region. However, computing 
the projection in 
many constrained %
settings is computationally prohibitive 
(\emph{e.g.}, 
projection over bounded 
trace norm matrices, flow polytope, matroid polytope, rotation matrices). In 
such scenarios, 
projection-free algorithms, \emph{a.k.a.}, Frank-Wolfe 
\citep{frank1956algorithm},  replace 
the projection with a linear program. Indeed, our proposed algorithm combines 
efficiently the 
zeroth-order information with solving a series of linear programs to ensure 
convergence to a 
near-optimal solution.

Continuous DR-submodular functions are an important subset of non-convex 
functions that can be 
minimized exactly \cite{bach2016submodular,staib2017robust}
and maximized approximately 
\cite{bian2017continuous,bian2017guaranteed,hassani2017gradient,
	mokhtari2018conditional,hassani2019stochastic,zhang2019one} %
This class of 
functions generalize the notion of diminishing returns, 
usually defined over discrete set functions, to the continuous domains. They 
have found 
numerous applications in machine learning including MAP inference in 
determinantal point processes (DPPs) 
\cite{kulesza2012determinantal}, experimental design 
\cite{chen2018online},
resource allocation \cite{eghbali2016designing}, 
mean-field inference in probabilistic models 
\cite{bian2018optimal}, 
among many others. 

\textbf{Motivation:}
\revise{Computing the gradient 
of a continuous DR-submodular function has been shown to be computationally prohibitive (or even 
intractable) in many applications. For example, 
the objective function of influence maximization is defined via specific  
stochastic processes \citep{kempe2003maximizing,rodriguez2012influence} and computing/estimating the gradient of the mutliliear extension would require a relatively high computational complexity. In the 
problem of D-optimal experimental design 
, the gradient of the objective function involves inversion of a potentially 
large matrix \citep{chen2018online}.} Moreover, when one 
attacks a submodular recommender model, only black-box information is available 
and the service provider is unlikely to provide additional first-order 
information (this is known as the black-box adversarial attack model) 
\citep{lei2019discrete}.

There has been very recent progress on developing zeroth-order methods for 
constrained optimization 
problems in convex and non-convex settings 
\cite{ghadimi2013stochastic,sahu2018towards}. Such methods typically assume 
the objective function is defined on the whole $\mathbb{R}^d$ so that they 
can sample points from a proper 
distribution defined on $\mathbb{R}^d$. For DR-submodular functions, this 
assumption might be unrealistic, since 
many DR-submodular 
functions might be only defined on a subset of $\mathbb{R}^d$, %
\emph{e.g.}, the multi-linear extension 
\cite{vondrak2008optimal}, a canonical example of DR-submodular functions, is 
only defined on a unit %
cube. Moreover, they can only guarantee to reach a first-order stationary 
point. However, 
\citet{hassani2017gradient} showed that for a monotone 
DR-submodular function, the stationary points can only guarantee $1/2$ 
approximation to the optimum. 
\revise{Therefore, if a state-of-the-art zeroth-order 
	non-convex algorithm is used for maximizing a monotone DR-submodular 
	function, it is likely to terminate at a suboptimal stationary point whose 
	approximation ratio is only $ 1/2 $.}

\textbf{Our contributions:} In this paper, we  propose a derivative-free 
and projection-free algorithm \Alg (BCG), that maximizes a monotone continuous 
DR-submodular 
function over 
a bounded convex body $\constraint\subseteq \mathbb{R}^d$. We 
consider three scenarios:

\begin{itemize}
\item In the deterministic setting, where function evaluations can be obtained 
exactly, BCG achieves the tight $[(1-1/e)OPT-\epsilon]$ approximation guarantee 
with $\mathcal{O}(d/\epsilon^3)$ function evaluations.%

\item In the stochastic setting, where function evaluations are noisy, BCG 
achieves the tight $[(1-1/e)OPT-\epsilon]$ approximation guarantee with 
$\mathcal{O}(d^3/\epsilon^5)$ function evaluations.

\item In the discrete setting, \Algdis (DBG) achieves the tight 
$[(1-1/e)OPT-\epsilon]$ 
approximation guarantee with $\mathcal{O}(d^5/\epsilon^5)$ function evaluations.
\end{itemize}

\begin{table*}[htb]%
	\centering
	\caption{Number of function queries in different 
		settings, where $D_1$ is the diameter of 
		$\constraint$.\label{tab:number_of_queries}}
	\begin{tabular}{lll}
		\toprule[1.5pt]
		Function & Additional Assumptions &Function Queries  \\ 
		\midrule
		continuous DR-submodular  & monotone, $G$-Lip., $L$-smooth 
		& 
		$\mathcal{O}(\max\{G, LD_1\}^3 \cdot \frac{d}{\epsilon^3})$ 
		[\cref{thm:zero}]\\ 
		stoch.\ conti.\ DR-submodular & monotone, $G$-Lip., 
		$L$-smooth &
		$\mathcal{O}(\max\{G, LD_1\}^3 \cdot \frac{d^3}{\epsilon^5})$ 
		[\cref{thm:zero_stochatic}] \\ 
		discrete submodular & monotone &
		$\mathcal{O}(\frac{d^5}{\epsilon^5})$ [\cref{thm:zero_discrete}]
		\\ 
		\bottomrule[1.25pt]
	\end{tabular}
\end{table*}

All the theoretical results are summarized in 
\cref{tab:number_of_queries}.

We would like to note that in discrete setting, due to  
the conservative upper bounds for the Lipschitz and smooth parameters of 
general multilinear extensions, %
and the variance of 
the gradient estimators 
subject to noisy function evaluations,
the required number of function 
queries in theory is larger than the best known result, 
$\mathcal{O}(d^{5/2}/\epsilon^3)$ 
in  \cite{mokhtari2018conditional,mokhtari2018stochastic}. However, our 
experiments show 
that empirically, our proposed algorithm often requires significantly fewer 
function evaluations and less running time, while achieving a practically 
similar utility.

\textbf{Novelty of our work:} All the previous results in 
constrained DR-submodular maximization assume access to (stochastic) gradients. 
In this work, we address a harder problem, \emph{i.e.}, we provide the first 
rigorous analysis when only (stochastic) function values can be obtained.
More specifically, with the smoothing trick \citep{flaxman2005online}, one can 
construct an unbiased gradient estimator via function queries. However, this 
estimator has a large $\mathcal{O}(d^2/\delta^2)$ variance which may cause  
FW-type 
methods to diverge. To overcome this issue, we build on the momentum method 
proposed by \citet{mokhtari2018conditional} in which they assumed access to the 
\emph{first-order} information.
	
Given a point $x$, the smoothed version of $F$ at $x$ is 
defined as 
$\expect_{v\sim B^d}[F(x+ \delta v)]$. If $x$ is close to the boundary of the 
domain $\domain, (x + \delta v)$ may fall outside of $\domain$, leaving the 
smoothed function undefined for many instances of DR-submodular functions 
(\emph{e.g.}, the multilinear extension is only defined over the unit cube). 
Thus the vanilla smoothing trick will not work. To this end, we transform the 
domain $\domain$ and constraint set $\constraint$ in a proper way 
and run our zeroth-order method on the transformed constraint set 
$\constraint'$. Importantly, we retrieve the same convergence rate of 
$\mathcal{O}(T^{-1/3})$ as in \cite{mokhtari2018conditional} with a minimum 
number of 
function queries in different settings (continuous, stochastic continuous, 
discrete).

We further note that by using more recent variance reduction techniques 
\citep{zhang2019one}, one might be able to reduce the required number of 
function evaluations.

\subsection{Further Related Work}
Submodular functions \cite{nemhauser1978analysis}, that capture the intuitive 
notion of diminishing returns, 
have become increasingly 
important in various machine learning applications. 
Examples include graph cuts in computer vision 
\cite{jegelka2011submodularity, jegelka2011approximation},  
data summarization \cite{lin2011word, lin2011class, 
tschiatschek2014learning,chen2018weakly,chen2017interactive},
influence maximization 
\cite{kempe2003maximizing,rodriguez2012influence,zhang2016influence},
feature compression \cite{bateni2019categorical}, 
network inference \cite{chen2017submodular},
active and semi-supervised 
learning 
\cite{guillory2010interactive, golovin2011adaptive, wei2015submodularity}, 
crowd 
teaching 
\cite{singla2014near}, 
dictionary 
learning \cite{das2011submodular}, fMRI parcellation 
\cite{salehi2017submodular}, compressed sensing and structured 
sparsity 
\cite{bach2010structured, bach2012optimization}, fairness in machine learning 
\cite{balkanski2015mechanisms, celis2016fair},  and learning causal structures 
\cite{steudel2010causal,zhou2016causal}, to name a few.
Continuous DR-submodular functions naturally extend the notion of 
diminishing returns to  the 
continuous domains 
\cite{bian2017guaranteed}. %
Monotone continuous 
DR-submodular functions 
can be (approximately) maximized over convex bodies using first-order methods 
\cite{bian2017guaranteed, 
	hassani2017gradient, mokhtari2018conditional}. 
Bandit maximization of monotone continuous DR-submodular 
functions~\cite{zhang2019online} is a closely related setting to ours.
However, to the best of 
our knowledge, none of the 
existing work has developed a 
zeroth-order 
algorithm for maximizing a monotone continuous DR-submodular function. For a 
detailed review of DFO and BBO, interested readers refer to book 
\citep{audet2017derivative}.

\section{Preliminaries}\label{sec:preliminaries}
\paragraph{Submodular Functions}
We say a set function $f:2^{\Omega}\to 
\mathbb{R}$ is submodular, if it satisfies the diminishing returns property: 
for any $A\subseteq B\subseteq \Omega$ and $x\in \Omega\setminus B$, we have 
\begin{equation}
f(A\cup\{x\})-f(A)\geq f(B\cup \{x\})-f(B).
\end{equation}
In words, the marginal gain of adding an element $x$ to a subset $A$ is no less 
than that of adding $x$ to its superset $B$.

For the continuous analogue, consider a function $F:\mathcal{X} \to 
\mathbb{R}_+$, where $\mathcal{X}=\Pi_{i=1}^n \mathcal{X}_i$, and each 
$\mathcal{X}_i$ is a compact subset of $\mathbb{R}_+$. We define $F$ to be 
continuous submodular if $F$ is continuous and for all $x, y \in \mathcal{X}$, 
we have 
\begin{equation}
F(x) + F(y) \geq F(x\vee y) + F(x \wedge y),
\end{equation}
where $\vee$ and $\wedge$ are the component-wise maximizing and minimizing 
operators, respectively.

The continuous function $F$ is  called DR-submodular \cite{bian2017guaranteed} 
if $F$ is differentiable and $\forall\, x \leq y: \,\nabla F(x) \geq \nabla 
F(y).$ 
An important implication of DR-submodularity is that the function $F$ is 
concave in any non-negative directions, \emph{i.e.}, for $x \leq y$, we have 
\begin{equation}
F(y) \leq F(x)+ \langle \nabla F(x), y-x\rangle.
\end{equation}
The function $F$ is called 
monotone if for $x \leq y$, we have
$F(x) \leq F(y).$    
\paragraph{Smoothing Trick}
For a function $F$ defined on $\mathbb{R}^d$, its $\delta$-smoothed version is 
given as %
\begin{equation}
\label{eq:smooth_definition}
\tF_\delta(x)\triangleq \expect_{v\sim B^d}[F(x+\delta v)],    
\end{equation}
where $v$ is chosen uniformly at random from the $d$-dimensional unit ball 
$B^d$. In words, the function $\tF_\delta$ at any point $x$ is obtained by 
``averaging''  $F$ over a ball of 
radius $\delta$ around $x$. In the sequel, we omit the subscript $\delta$ for 
the sake of simplicity and use  $\tF$ instead of $\tF_\delta$.

\cref{lem:smooth_approx} below shows that under the Lipschitz assumption for $F$, the 
smoothed 
version $\tF$ is a good approximation of $F$, and also inherits the key 
structural properties of  
$F$ (such as monotonicity and submodularity). Thus one can 
(approximately) optimize $F$ via optimizing~$\tF$. 

\begin{lemma}[Proof in 
	\cref{app:proof_lemma_smooth_approx}]\label{lem:smooth_approx}
	If $F$ is monotone continuous 
	DR-submodular and $G$-Lipschitz continuous on $\mathbb{R}^d$,  %
	then %
	so is $\tF$ and 
	\begin{equation}
	|\tF(x)-F(x)|\le \delta G.
	\end{equation}   
\end{lemma}

An important property of $\tF$ is that one can obtain an unbiased estimation for its 
gradient 
$\nabla \tF$ by a single query of $F$. This property plays a key 
role in our proposed derivative-free algorithms.

\begin{lemma}[Lemma 6.5 in \citep{hazan2016introduction}]
	\label{lem:gradient_of_smooth} 
	Given a function $F$ on $\mathbb{R}^d$, if we choose $u$ uniformly at 
	random from the $(d-1)$-dimensional unit 
	sphere $S^{d-1}$, then %
	we have 
	\begin{equation} \label{eq:gradient_smooth}
	\nabla \tF(x)= \expect_{u \sim S^{d-1}}\left[\frac{d}{\delta}F(x+\delta 
	u)u\right].    
	\end{equation}
\end{lemma}

\section{DR-Submodular Maximization}
In this paper, we mainly focus on the constrained optimization 
problem: 
\begin{equation}
\max_{x \in \constraint} F(x),
\end{equation}
where $F$ is a monotone continuous DR-submodular function on $\mathbb{R}^d$, and
the constraint set $\constraint \subseteq \mathcal{X} \subseteq\mathbb{R}^d$ is 
convex and compact. %

For \emph{first-order} monotone DR-submodular maximization, one can use \AlgCG 
\cite{calinescu2011maximizing, bian2017guaranteed}, a variant of Frank-Wolfe 
Algorithm \citep{frank1956algorithm,jaggi2013revisiting,lacoste2015global}, 
to achieve the  $[(1-1/e)OPT-\epsilon]$ approximation 
guarantee. At iteration $t$, the FW variant first maximizes the linearization 
of the objective function $F$: 
\begin{equation}
v_t = \argmax_{v \in \constraint} \langle v, 
\nabla F(x_t) \rangle.
\end{equation}
Then the current point $x_t$ moves 
in the direction of $v_t$ with a step size $\gamma_t \in (0,1]$: 
\begin{equation}
x_{t+1} = x_t + \gamma_t v_t.
\end{equation}
Hence, by solving linear optimization problems, the iterates 
are updated without resorting to the projection oracle. 

Here we introduce our main algorithm \Alg which assumes access only 
to function values (\emph{i.e.}, zeroth-order information). This algorithm is 
partially based on the idea of \AlgCG. The basic idea is to utilize the 
function evaluations of $F$ at carefully selected points to obtain unbiased 
estimations of the gradient of the smoothed version, $\nabla \tF$. By 
extending %
\AlgCG to the derivative-free setting %
and using recently proposed variance reduction 
techniques, we can then optimize $\tF$ 
near-optimally. Finally, by \cref{lem:smooth_approx} we show that the obtained 
optimizer also provides a good solution for $F$. 
Recall that continuous DR-submodular functions are 
defined on a box $\mathcal{X}=\Pi_{i=1}^n \mathcal{X}_i$. To simplify the exposition, 
we can assume, without loss of generality, that
the  objective function $F$ is defined on $\domain\triangleq \prod_{i=1}^d 
[0,a_i]$ \cite{bian2017continuous}.
Moreover, we note that %
since $\tF = \expect_{v\sim B^d}[F(x+\delta v)]$, for $x$ close to $\partial 
\domain$ (the boundary 
of $\domain$), the point 
$x + \delta v$ may fall outside of $\domain$, leaving the function $\tF$ 
undefined. 

To circumvent this issue, we shrink the 
domain $\domain$ by $\delta$. Precisely, the shrunk domain is defined as
\begin{equation} \label{D'}
\domain'_\delta = \{x \in \domain| d(x, \partial \domain) \geq \delta \}.
\end{equation}
Since we assume  
$\domain = \prod_{i=1}^d [0,a_i]$, the shrunk domain is 
$\domain'_\delta = 
\prod_{i=1}^d[\delta, a_i-\delta]$.
Then for all $x \in \domain'_\delta$, we have $x 
+ \delta v \in \domain$.  So $\tF$ is well-defined on $\domain'_\delta$.
By \cref{lem:smooth_approx}, the optimum of 
$\tF$ on the shrunk domain $\domain'_\delta$ will be 
close to that on the original domain $\domain$, if $\delta$ is small enough. 
Therefore, we can first optimize $\tF$ on $\domain'_\delta$, then
approximately optimize $\tF$ (and thus $F$) on $\domain$. For simplicity of 
analysis, we also translate the shrunk domain 
$\domain'_\delta$
by $-\delta$, and denote it as $\domainsh = \prod_{i=1}^d [0, a_i-2\delta]$. 

Besides the domain $\domain$, 
we also need to consider the transformation on constraint set 
$\constraint$. Intuitively, if there is no translation, we should consider the 
intersection of $\constraint$ and the shrunk domain $\domain'_\delta$. But 
since we translate $\domain'_\delta$ by $-\delta$, the same transformation 
should be performed on $\constraint$. Thus, we define the transformed 
constraint set as the translated intersection (by $-\delta$) of 
$\domain'_\delta$ and $\constraint$:
\begin{equation} \label{K'}
\constraint'\triangleq  (\domain'_\delta \cap \constraint) - \delta \one 
=\domainsh \cap (\constraint-\delta\one).
\end{equation} 

It is well known 
that the FW Algorithm is  sensitive to 
the accuracy of gradient, and may have arbitrarily poor performance with 
stochastic gradients \cite{hazan2016variance,mokhtari2018stochastic}. Thus we 
incorporate 
two methods of variance reduction into our 
proposed algorithm \Alg which 
correspond to Step 7 and Step 8 in \cref{alg:zero_frank_wolfe}, respectively. 
First, instead of the one-point gradient estimation in 
\cref{lem:gradient_of_smooth}, we adopt the two-point estimator of $\nabla 
\tF(x)$
\citep{agarwal2010optimal,shamir2017optimal}:
\begin{equation}
\label{eq:gradient_smooth_two_points}
\frac{d}{2\delta}(F(x+\delta u) - F(x-\delta u)) u,
\end{equation}
where $u$ is chosen uniformly at random from the unit sphere $S^{d-1}$.We note 
that \eqref{eq:gradient_smooth_two_points} is an unbiased gradient estimator  
with less variance w.r.t. the one-point estimator. We also average over a 
mini-batch of $B_t$ independently sampled two-point estimators for further 
variance reduction. The second 
variance-reduction 
technique is the momentum method used in 
\citep{mokhtari2018conditional} to estimate the gradient by a vector 
$\bar{g}_t$ which is updated at each iteration as follows:
\begin{equation}
\label{eq:averaging}
\bar{g}_t = (1-\rho_t) \bar{g}_{t-1} + \rho_t g_t.
\end{equation}
Here $\rho_t$ is a given step size, $\bar{g}_0$ is initialized as an all zero vector $\mathbf{0}$, and $g_t$ 
is an unbiased estimate of the gradient at iterate $x_t$. As $\bar{g}_t$ is a 
weighted average of previous gradient approximation $\bar{g}_{t-1}$ and the 
newly updated stochastic gradient $g_t$, it has a lower variance compared with 
$g_t$. Although $\bar{g}_t$ is not an unbiased estimation of the true gradient, 
the error of it will approach zero as time proceeds. The detailed description 
of \Alg is provided in \cref{alg:zero_frank_wolfe}. 

\begin{algorithm}[t!]
	\begin{algorithmic}[1]
		\STATE {\bfseries Input:} constraint set $\constraint$, iteration 
		number $T$, radius $\delta$, step size $\rho_t$, batch size $B_t$
		\STATE {\bfseries Output:} $x_{T+1}+\delta \one$
		\STATE $x_1\gets \mathbf{0}, \enspace \bar{g}_0 \gets \mathbf{0}$
		\FOR{$t= 1$ {\bfseries to} $T$}
		\STATE Sample $u_{t,1},\dots, u_{t,B_t}$ i.i.d.\ from $S^{d-1}$
		\STATE For $i=1$ to $B_t$, let $y_{t,i}^+ \gets \delta \one+x_t+\delta 
		u_{t,i}, y_{t,i}^- \gets \delta \one+x_t-\delta u_{t,i}$ and evaluate 
		$F(y_{t,i}^+), F(y_{t,i}^-)$ 
		\STATE $g_t\gets \frac{1}{B_t}\sum_{i=1}^{B_t} 
		\frac{d}{2\delta}[F(y_{t,i}^+)-F(y_{t,i}^-)]u_{t,i}$
		\STATE $\bar{g}_t\gets (1-\rho_t)\bar{g}_{t-1}+\rho_t g_t$
		\STATE $v_t\gets \argmax_{v\in \constraint' } \langle 
		v,\bar{g}_t\rangle $
		\STATE $x_{t+1}\gets x_t+\frac{v_t}{T}$
		\ENDFOR
		\STATE Output $x_{T+1}+\delta \one$
	\end{algorithmic}
	\caption{\Alg}\label{alg:zero_frank_wolfe}
\end{algorithm} 

\begin{mdframed}
\begin{theorem}[Proof in \cref{app:theorem_zero}]
	\label{thm:zero}
	For a monotone continuous DR-submodular function $F$, which is also 
	$G$-Lipschitz continuous and $L$-smooth on a 
	convex and compact constraint set $\constraint$, if we set 
	$\rho_t=2/(t+3)^{2/3}$ in \cref{alg:zero_frank_wolfe}, then we have
	\begin{align*}
	&(1-1/e)F(x^*)-\expect[F(x_{T+1}+\delta\one)] \\
	\le{}&  
	\frac{3D_1Q^{1/2}}{T^{1/3}}+ \frac{LD_1^2}{2 T} 
	+ \delta G(1+(\sqrt{d}+1)(1-1/e)).
	\end{align*}
	where $Q= \max \{ 4^{2/3}G^2, 
	4cdG^2/B_t+ 6L^2D_1^2 \}, c$ is a constant, $D_1= \diam(\constraint')$, and 
	$x^*$ is the global 
	maximizer of $F$ on $\constraint$.
\end{theorem}
\end{mdframed}
\begin{remark}
By setting $T=\mathcal{O}(1/\epsilon^3)$, $B_t=d$, and 
$\delta=\epsilon/\sqrt{d}$, the error term (RHS) is guaranteed to be at 
most $\mathcal{O}(\epsilon)$. Also, the total number of function evaluations is 
at 
most $\mathcal{O}(d/\epsilon^3)$.	
\end{remark}

We can also extend \cref{alg:zero_frank_wolfe} to the stochastic case in which
we obtain information 
about $F$ only through its noisy function evaluations $\hat{F}(x)=F(x)+\xi$, 
where $\xi$ is stochastic zero-mean noise. In particular,   
in Step 6 of \cref{alg:zero_frank_wolfe}, we obtain independent stochastic 
function evaluations $\hat{F}(y_{t,i}^+)$ and $ \hat{F}(y_{t,i}^-)$, instead of the 
exact function values $F(y_{t,i}^+)$ and $F(y_{t,i}^-)$. For unbiased function 
evaluation oracles with uniformly bounded variance, we have the following 
theorem.  
\begin{mdframed}
\begin{theorem}[Proof in \cref{app:theorem_zero_stochastic}]
	\label{thm:zero_stochatic}
	Under the condition of \cref{thm:zero}, if we further assume that for all 
	$x$, $\expect[\hat{F}(x)]=F(x)$ and $\expect[|\hat{F}(x)-F(x)|^2]\leq 
	\sigma_0^2$, then we have
	\begin{align*}
	&(1-1/e)F(x^*)-\expect[F(x_{T+1}+\delta\one)] \\
	\le{}&  
	\frac{3D_1Q^{1/2}}{T^{1/3}}+ \frac{LD_1^2}{2 T} 
	+ \delta G(1+(\sqrt{d}+1)(1-1/e)),
	\end{align*}
	where $D_1= \diam(\constraint'), Q= \max \{4^{2/3}G^2, 
	6L^2D_1^2+(4cdG^2+2d^2\sigma_0^2/\delta^2)/B_t\}, c$ is a constant, and
	$x^*$ is the global 
	maximizer of $F$ on $\constraint$.
\end{theorem}
\end{mdframed}
\begin{remark}
By setting $T=\mathcal{O}(1/\epsilon^3)$, 
$B_t=d^3/\epsilon^2$, and $\delta=\epsilon/\sqrt{d}$, 
the error term (RHS) is at most $\mathcal{O}(\epsilon)$.
The total number of evaluations is at most 
$\mathcal{O}(d^3/\epsilon^5)$.
\end{remark}
\section{Discrete Submodular Maximization}\label{sec:discrete}
In this section, we describe how \Alg can be used to solve a discrete 
submodular maximization problem 
with a general matroid constraint, \emph{i.e.}, $\max_{S \in \mathcal{I}} 
f(S)$, 
where $f$ is a monotone submodular set function and $\mathcal{I}$ is a 
matroid.

For any monotone submodular set function $f: 2^\Omega \to \mathbb{R}_{\geq 0}$, 
its multilinear extension $F:[0,1]^d \to \mathbb{R}_{\geq 0}$, defined as 
\begin{equation}
F(x)=\sum_{S \subseteq \Omega}f(S)\prod_{i \in S}x_i\prod_{j \notin S}(1-x_j),
\end{equation}
is monotone and DR-submodular \citep{calinescu2011maximizing}. Here, 
$d=|\Omega|$ is
the size of the ground set $\Omega$. %
Equivalently, 
we have 
$F(x) = \expect_{S \sim x}[f(S)],$ where  $S \sim x$ means that the each 
element $i \in \Omega$ is included in $S$ with probability $x_i$ independently.

It can be shown that in lieu of solving the discrete optimization problem 
one can solve the continuous optimization problem $\max_{x \in \mathcal{K}} 
F(x),$
where %
$\mathcal{K} = \text{conv}\{1_I: I 
\in \mathcal{I}\}$ is the matroid polytope \citep{calinescu2011maximizing}. 
This equivalence is 
obtained by showing that (i) the optimal values of the two problems 
are the same, and (ii) for any fractional 
vector $x \in \mathcal{K}$ we can deploy efficient, lossless rounding 
procedures that produce a set $S \in \mathcal{I}$ such that $\mathbb{E}[f(S)] 
\geq F(x)$ (\emph{e.g.}, pipage 
rounding \citep{ageev2004pipage,calinescu2011maximizing} and contention 
resolution \citep{chekuri2014submodular}). So we can view $\tF$ as the 
underlying function that we intend to optimize, and invoke \Alg. 
As a result, we want that $F$ is $G$-Lipschitz 
and $L$-smooth as in \cref{thm:zero}. The following lemma shows these 
properties are 
satisfied automatically %
if $f$ is bounded. 
\begin{lemma}
	\label{lem:discrete_to_continuous}
	For a submodular set function $f$ defined on $\Omega$ with $\sup_{X 
		\subseteq \Omega} |f(X)| 
	\le M$, its multilinear extension $F$ is $2M\sqrt{d}$-Lipschitz and  
	$4M\sqrt{d(d-1)}$-smooth.      
\end{lemma}

We note that the bounds for Lipschitz and smoothness parameters 
actually 
depend on the norms that we consider. However, different norms are equivalent 
up 
to a factor that may depend on the dimension. If we consider another norm, some 
dimension factors may be absorbed into the norm. Therefore, we only study the 
Euclidean norm in \cref{lem:discrete_to_continuous}.

We further note that computing the exact value of %
$F$ is difficult as it requires evaluating $f$ over all the subsets $S \in 
\Omega$.  However, one can construct an unbiased estimate for the value $F(x)$  
by simply sampling a random set $S\sim x$ and returning $f(S)$ as the estimate. 
We present our algorithm in detail in 
\cref{alg:zero_frank_wolfe_discrete},  
where we have $\domain = [0, 1]^d$, since 
$F$ is defined on $[0, 1]^d$, and thus $\domainsh = [0, 
1-2\delta]^d$. %
We state the theoretical result formally in \cref{thm:zero_discrete}.

\begin{algorithm}[t!]
	\begin{algorithmic}[1]
		\STATE {\bfseries Input:} matroid constraint $\matroid$, transformed 
		constraint set $\constraint' = \domainsh\cap (\constraint - \delta 
		\one)$ where $\constraint = \text{conv}\{1_I: I \in \mathcal{I}\}$,	
		number of iterations $T$, radius $\delta$, step size $\rho_t$, batch 
		size $B_t$, sample size $S_{t,i}$	
		\STATE {\bfseries Output:} $X_{T+1}$
		\STATE $x_1\gets \mathbf{0}, \enspace \bar{g}_0 \gets \mathbf{0}$, 
		\FOR{$t=1$ {\bfseries to} $T$}
		\STATE Sample $u_{t,1},\dots, u_{t,B_t}$ i.i.d.\ from $S^{d-1}$
		\STATE For $i=1$ to $B_t$, let $y_{t,i}^+ \gets \delta \one+x_t+\delta 
		u_{t,i}, 
		y_{t,i}^- \gets \delta \one+x_t-\delta u_{t,i}$, independently sample 
		subsets 
		$Y_{t,i}^+$ and $Y_{t,i}^-$ for $S_{t,i}$ times according to 
		$y_{t,i}^+, 
		y_{t,i}^-$, get sampled subsets $Y_{t,i,j}^+, Y_{t,i,j}^-, \forall j 
		\in 
		[S_{t,i}]$, evaluate the function values 
		$f(Y_{t,i,j}^+), f(Y_{t,i,j}^-), \forall j \in 
		[S_{t,i}]$, and calculate the averages 
		$\bar{f}_{t,i}^+ \gets \frac{\sum_{j=1}^{S_{t,i}} 
		f(Y_{t,i,j}^+)}{S_{t,i}}, 
		\bar{f}_{t,i}^- \gets \frac{\sum_{j=1}^{S_{t,i}} 
		f(Y_{t,i,j}^-)}{S_{t,i}}$ 
		\STATE $g_t\gets \frac{1}{B_t}\sum_{i=1}^{B_t} 
		\frac{d}{2\delta}(\bar{f}_{t,i}^+-\bar{f}_{t,i}^-)u_{t,i}$
		\STATE $\bar{g}_t\gets (1-\rho_t)\bar{g}_{t-1}+\rho_t g_t$
		\STATE $v_t\gets \argmax_{v\in \constraint' } \langle 
		v,\bar{g}_t\rangle $
		\STATE $x_{t+1}\gets x_t+\frac{v_t}{T}$
		\ENDFOR
		\STATE Output $X_{T+1} = \text{round}(x_{T+1}+\delta \one)$
	\end{algorithmic}
	\caption{\Algdis}\label{alg:zero_frank_wolfe_discrete}
\end{algorithm}

\begin{mdframed}
\begin{theorem}[Proof in \cref{app:theorem_discrete}]\label{thm:zero_discrete}
	For a monotone submodular set function $f$ with $\sup_{X \subseteq \Omega} 
|f(X)| \le M$, if we set $\rho_t=2/(t+3)^{2/3}, S_{t,i}=l$ in 
\cref{alg:zero_frank_wolfe_discrete}, then we have	
\begin{align*}
&(1-1/e)f(X^*) - 
\expect[f(X_{T+1})]\\
\le{} & 
\frac{3D_1Q^{1/2}}{T^{1/3}}+ \frac{2M\sqrt{d(d-1)}D_1^2}{T} 
\\
& \quad + 2M\delta \sqrt{d}(1+(\sqrt{d}+1)(1-1/e)).
\end{align*}
where $D_1= \diam(\constraint')$, $Q= \max 
\{\frac{2d^2M^2(\frac{1}{l\delta^2}+8c)}{B_t}+ 96d(d-1)M^2D_1^2, 
4^{5/3}dM^2 \}, c$ is a constant, $X^*$ is the global maximizer of $f$ 
under matroid constraint $\matroid$.	
\end{theorem}
\end{mdframed}
\begin{remark}
	By setting $T=\mathcal{O}(d^3/\epsilon^3)$, $B_t=1, l =d^2/\epsilon^2$, and 
	$\delta=\epsilon/d$, the error term (RHS) is at most 
	$\mathcal{O}(\epsilon)$. 
	The total number of evaluations is at most $\mathcal{O}(d^5/\epsilon^5)$.
\end{remark} 

We note that in \cref{alg:zero_frank_wolfe_discrete}, 
$\bar{f}^+_{t,i}$ is the unbiased estimation of $F(y^+_{t,i})$, and the same 
holds for $\bar{f}^-_{t,i}$ and $F(y^-_{t,i})$. As a result, we can 
analyze the algorithm under the framework of stochastic continuous 
submodular maximization. By applying \cref{thm:zero_stochatic}, 
\cref{lem:discrete_to_continuous}, and the facts $\expect[|\bar{f}^+_{t,i}- 
F(y^+_{t,i})|^2] \le M^2/S_{t,i}, \expect[|\bar{f}^-_{t,i}- 
F(y^-_{t,i})|^2] \le M^2/S_{t,i}$ directly, we can also attain 
\cref{thm:zero_discrete}.
\section{Experiments}

\newcommand{\AlgSCG}{\textsf{Stochastic Continuous Greedy}\xspace}
\newcommand{\AlgZGA}{\textsf{Zeroth-Order Gradient Ascent}\xspace}
\newcommand{\AlgGA}{\textsf{Gradient Ascent}\xspace}

In this section, we will compare \Alg (BCG) and \Algdis (DBG) with the following
baselines: 

\begin{itemize}
\item \AlgZGA (ZGA) is the projected gradient ascent algorithm 
equipped with the same two-point gradient estimator as BCG uses. Therefore, it 
is a \emph{zeroth-order} projected algorithm. 

\item \AlgSCG (SCG) is the 
state-of-the-art \emph{first-order} algorithm for maximizing continuous 
DR-submodular functions~\cite{mokhtari2018conditional,mokhtari2018stochastic}. 
Note that it is a projection-free algorithm. 

\item \AlgGA (GA) is the 
\emph{first-order} projected gradient ascent 
algorithm~\cite{hassani2017gradient}.
\end{itemize}

The stopping criterion for the algorithms is whenever a given 
number of iterations 
is achieved. Moreover, the batch sizes
$S_{t,i}$ in \cref{alg:zero_frank_wolfe} and $B_t$ in 
\cref{alg:zero_frank_wolfe_discrete} are both 1. Therefore, in the experiments, 
DBG uses 1 query per iteration while SCG uses $\mathcal{O}(d)$ queries.

\begin{figure*}[tb]
	\centering
	\begin{subfigure}[t]{0.24\textwidth}
		\includegraphics[width=\textwidth]{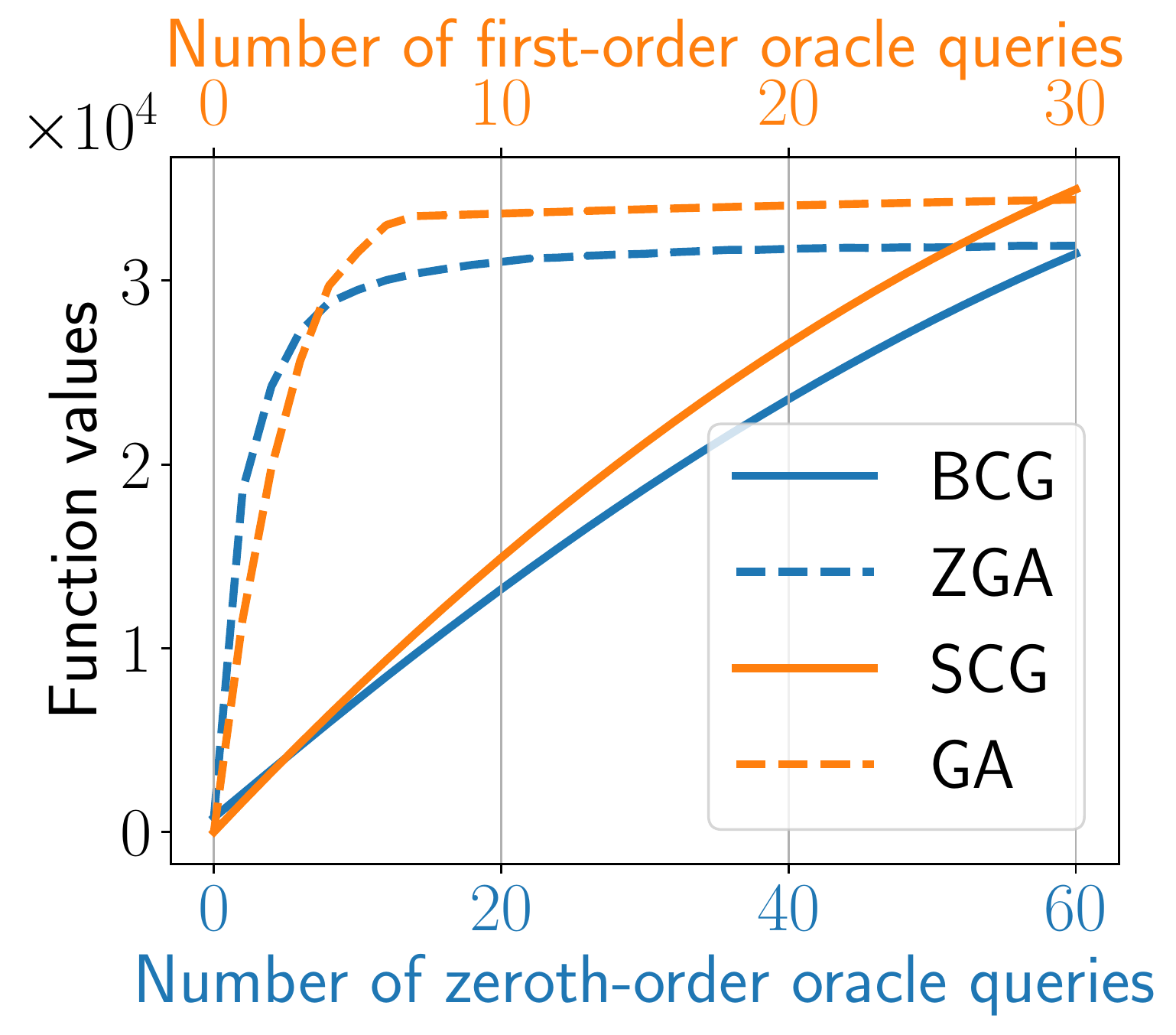}
		\caption{NQP}
		\label{fig:nqp_function}
	\end{subfigure}
	\begin{subfigure}[t]{0.24\textwidth}
		\includegraphics[width=\textwidth]{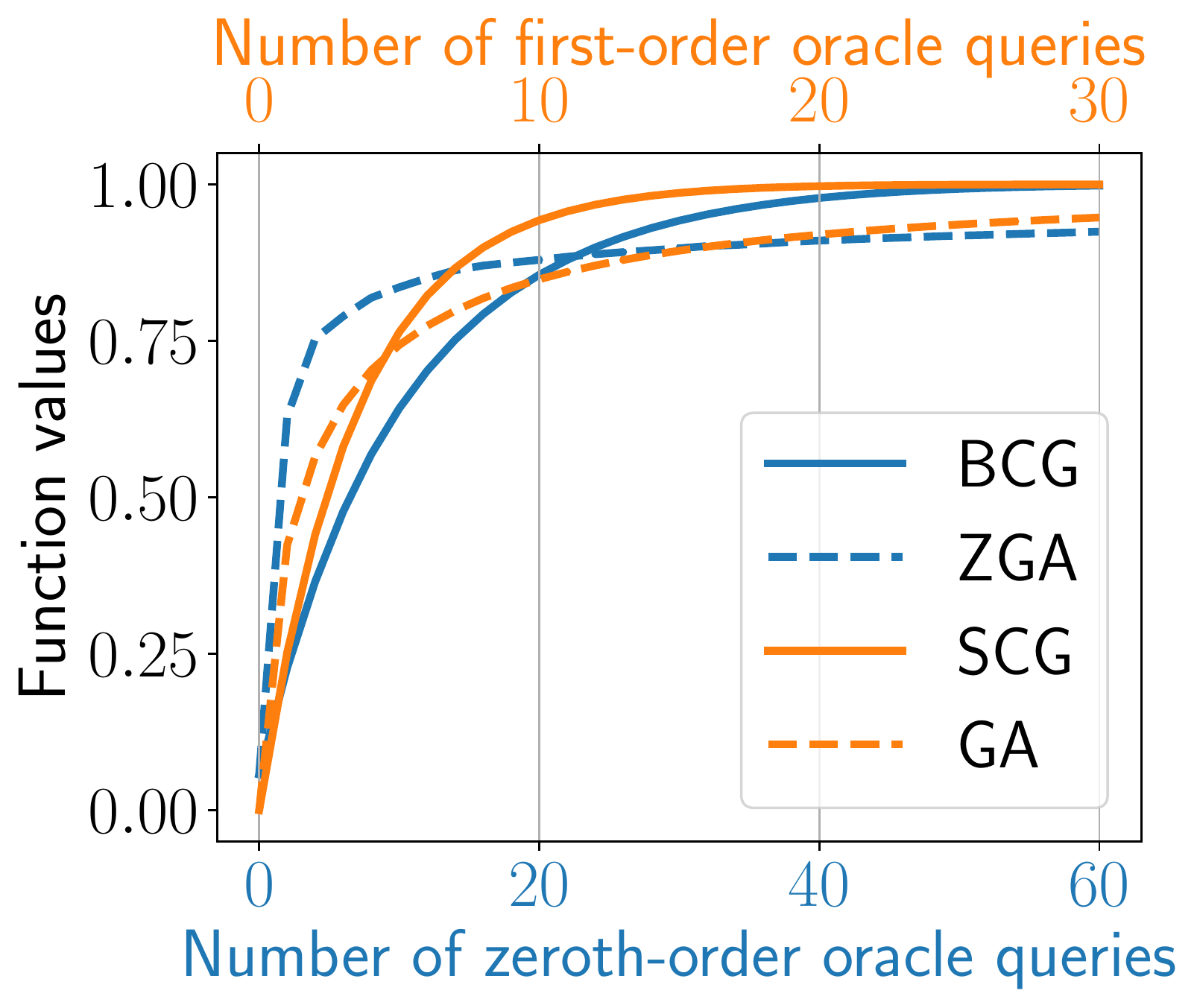}
		\caption{Topic summarization}
		\label{fig:topic_function}
	\end{subfigure}
	\begin{subfigure}[t]{0.24\textwidth}
		\includegraphics[width=\textwidth]{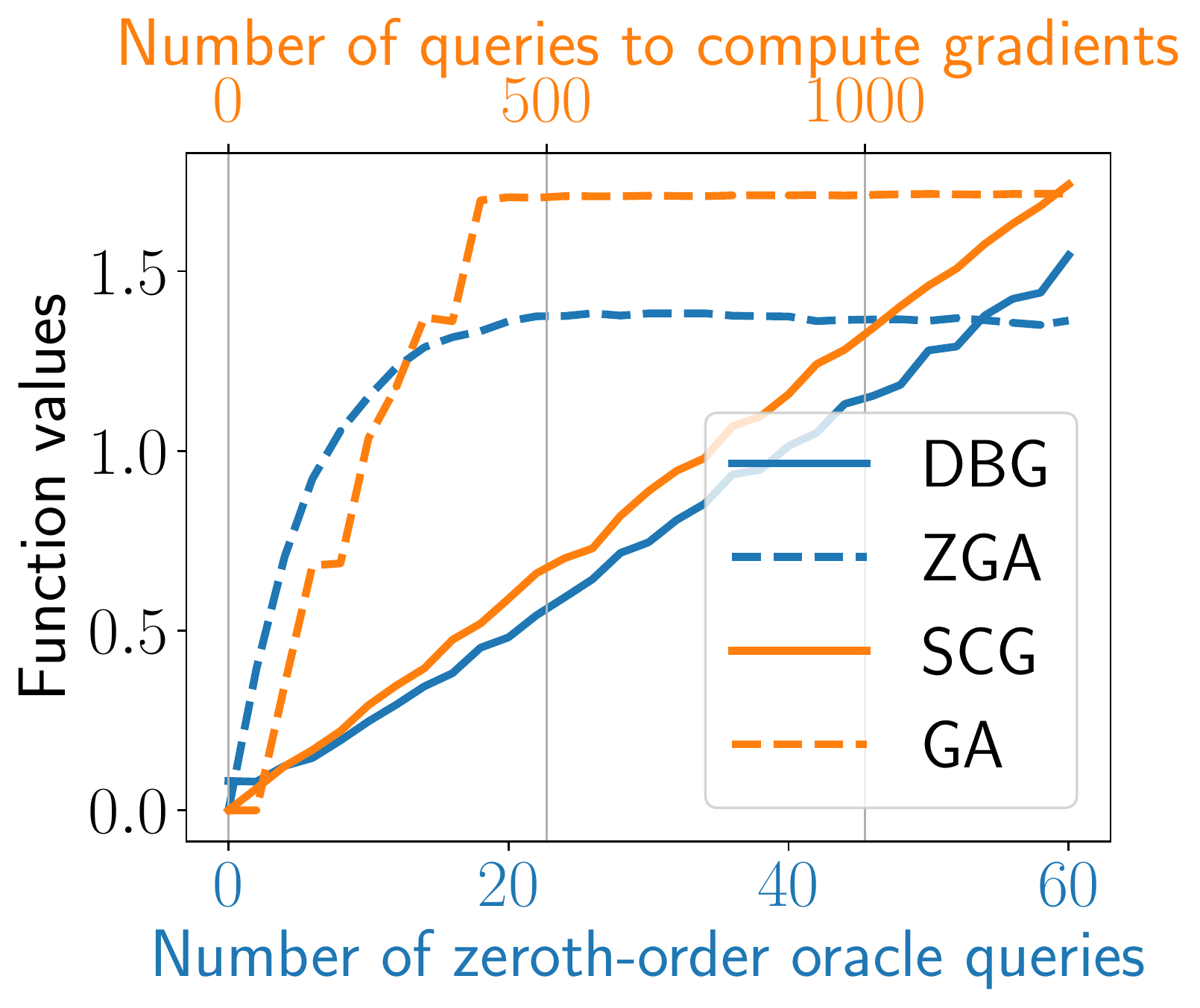}
		\caption{Active set selection}
		\label{fig:active_function}
	\end{subfigure}
	\begin{subfigure}[t]{0.24\textwidth}
		\includegraphics[width=\textwidth]{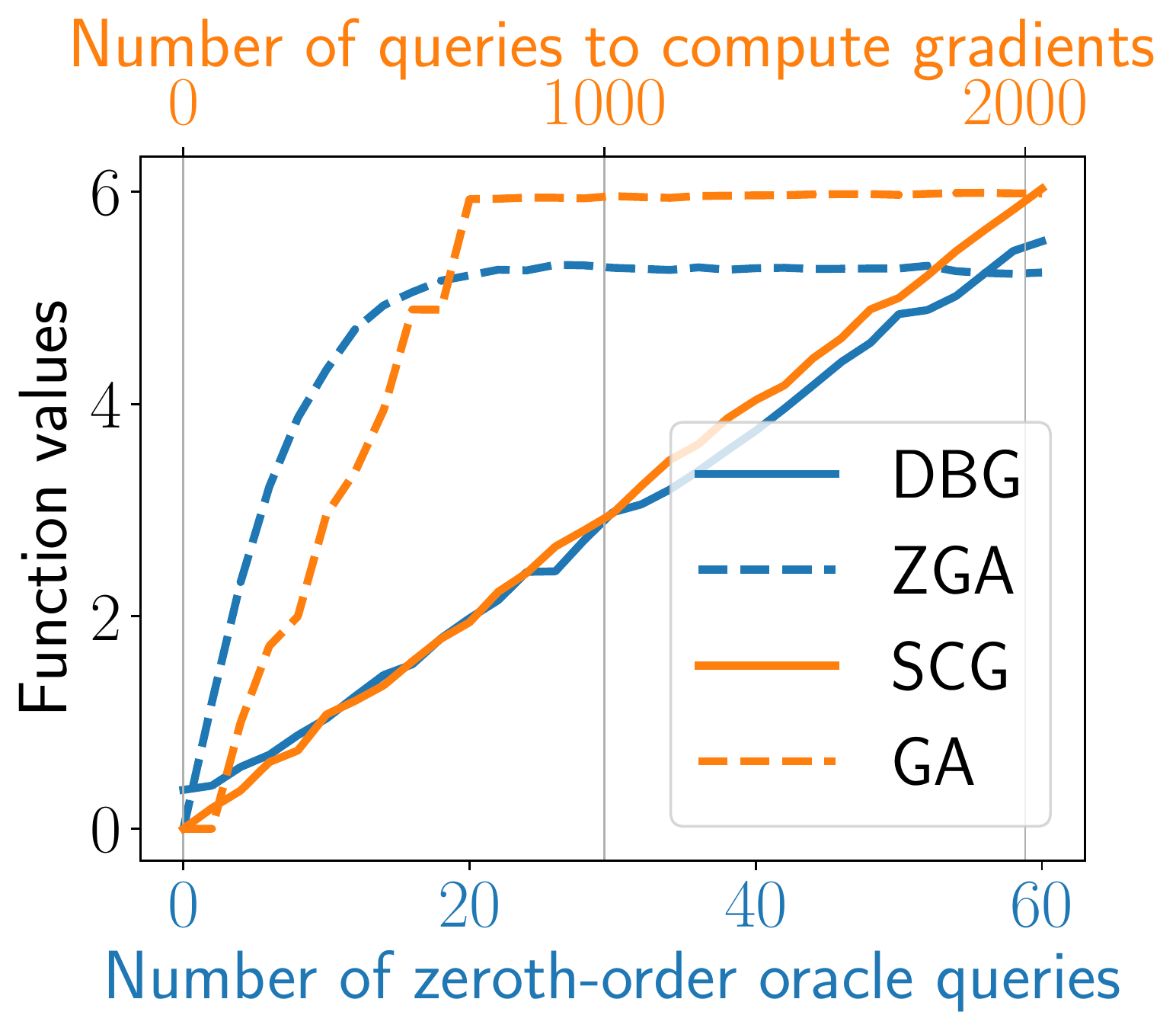}
		\caption{Influence maximization}
		\label{fig:cover_function}
	\end{subfigure}
	\caption{Function value vs.\ number of oracle queries. Note that every 
		chart has dual horizontal axes. 
		\textcolor{orange}{Orange} lines use the \textcolor{orange}{orange} 
		horizontal axes above while \textcolor{RoyalBlue}{blue} lines use the 
		\textcolor{RoyalBlue}{blue} ones below. }\label{fig:function}
\end{figure*}

\begin{figure*}[htb]\label{fig:time}
	\centering
	\begin{subfigure}[t]{0.23\textwidth}
		\includegraphics[width=\textwidth]{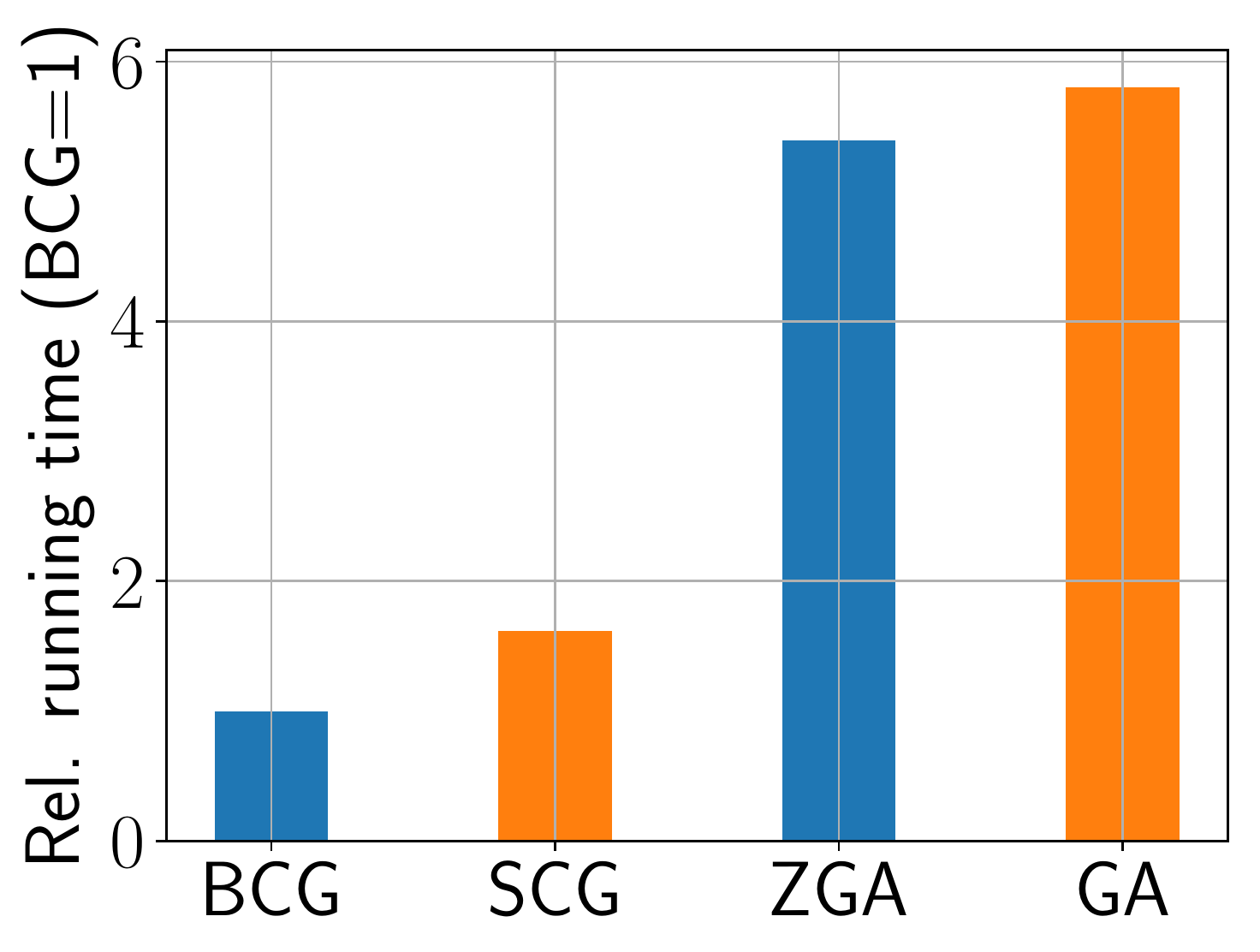}
		\caption{NQP}
		\label{fig:nqp_time}
	\end{subfigure}
	\begin{subfigure}[t]{0.23\textwidth}
		\includegraphics[width=\textwidth]{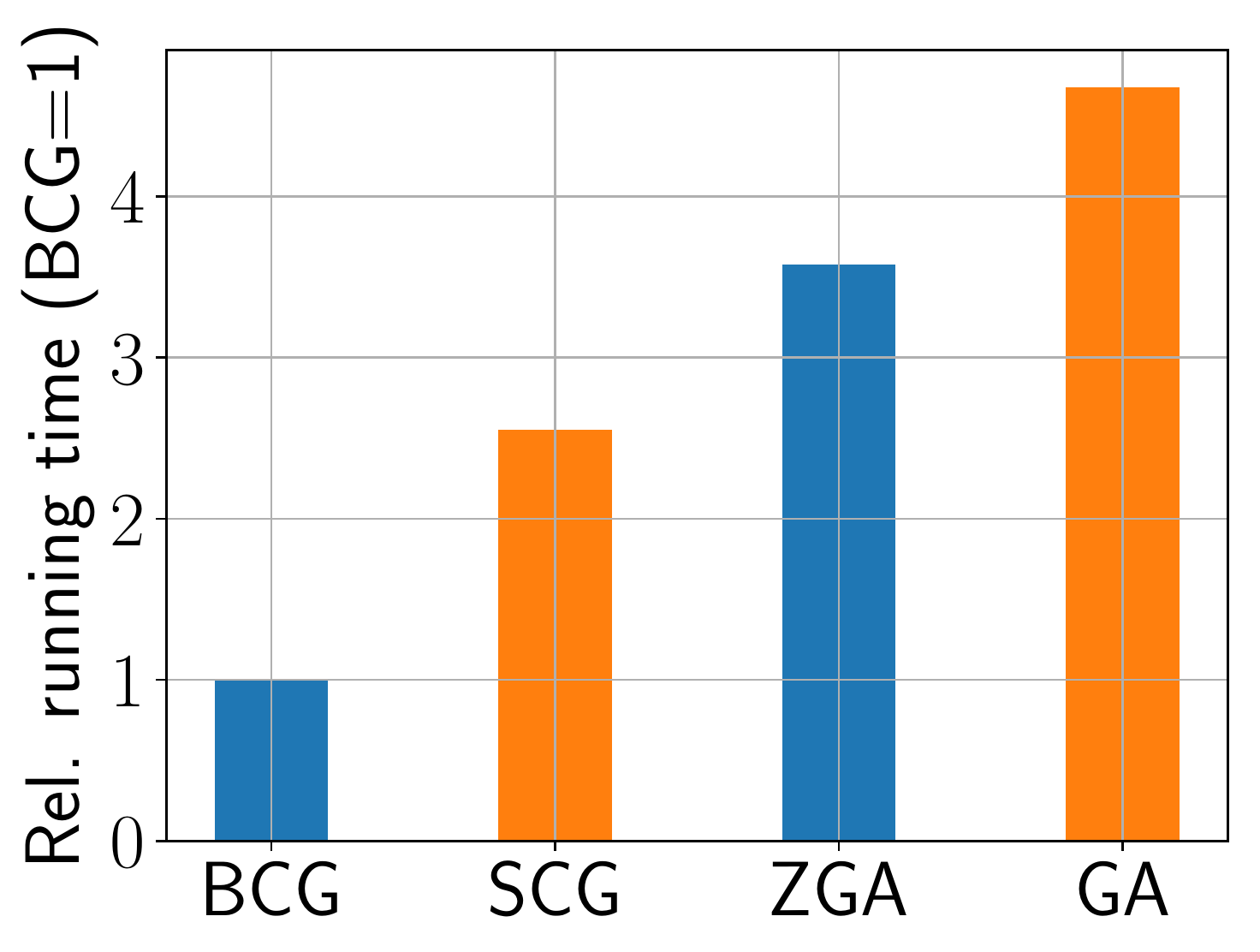}
		\caption{Topic summarization}
		\label{fig:topic_time}
	\end{subfigure}
	\begin{subfigure}[t]{0.245\textwidth}
		\includegraphics[width=\textwidth]{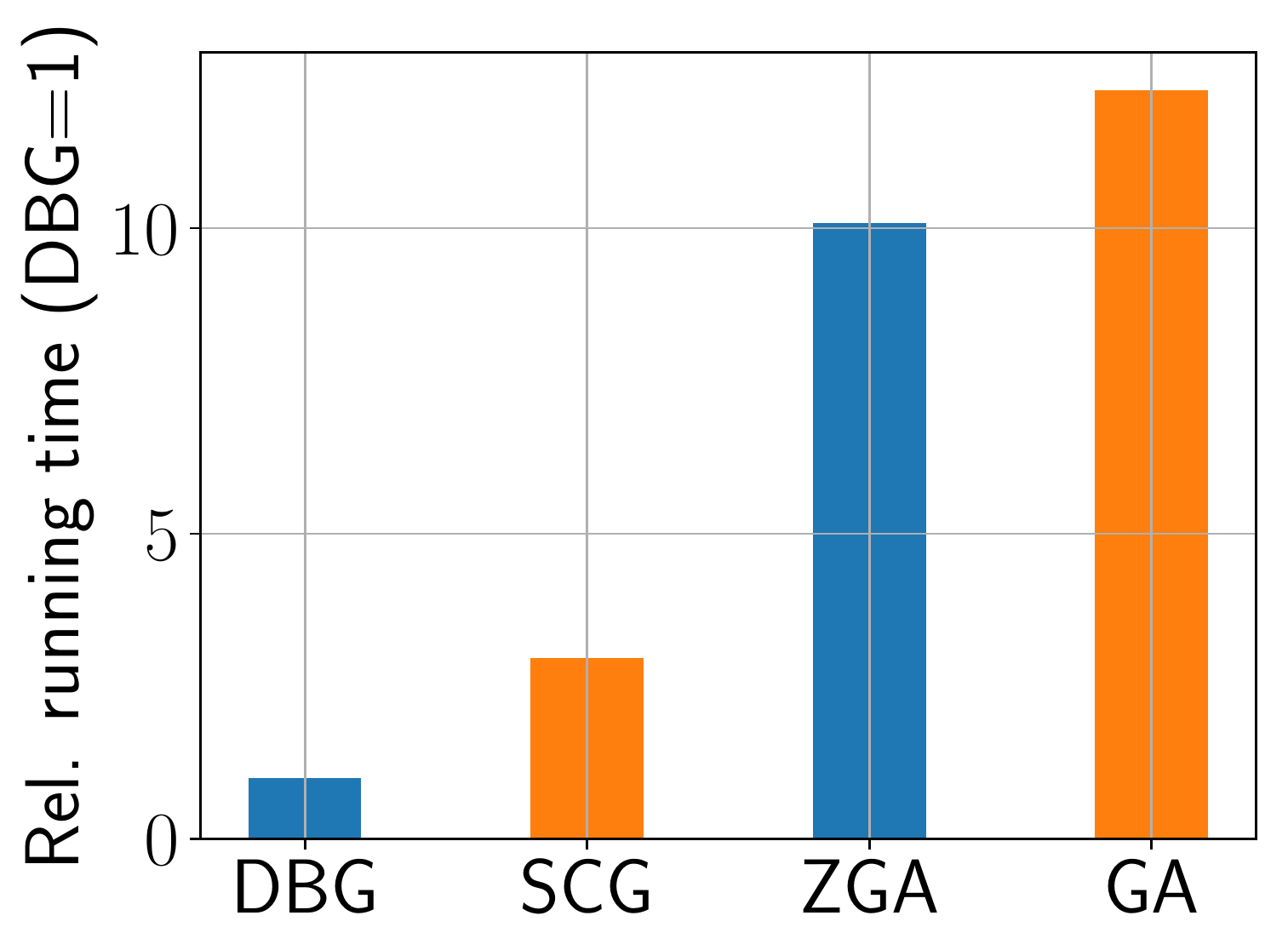}
		\caption{Active set selection}
		\label{fig:active_time}
	\end{subfigure}
	\begin{subfigure}[t]{0.23\textwidth}
		\includegraphics[width=\textwidth]{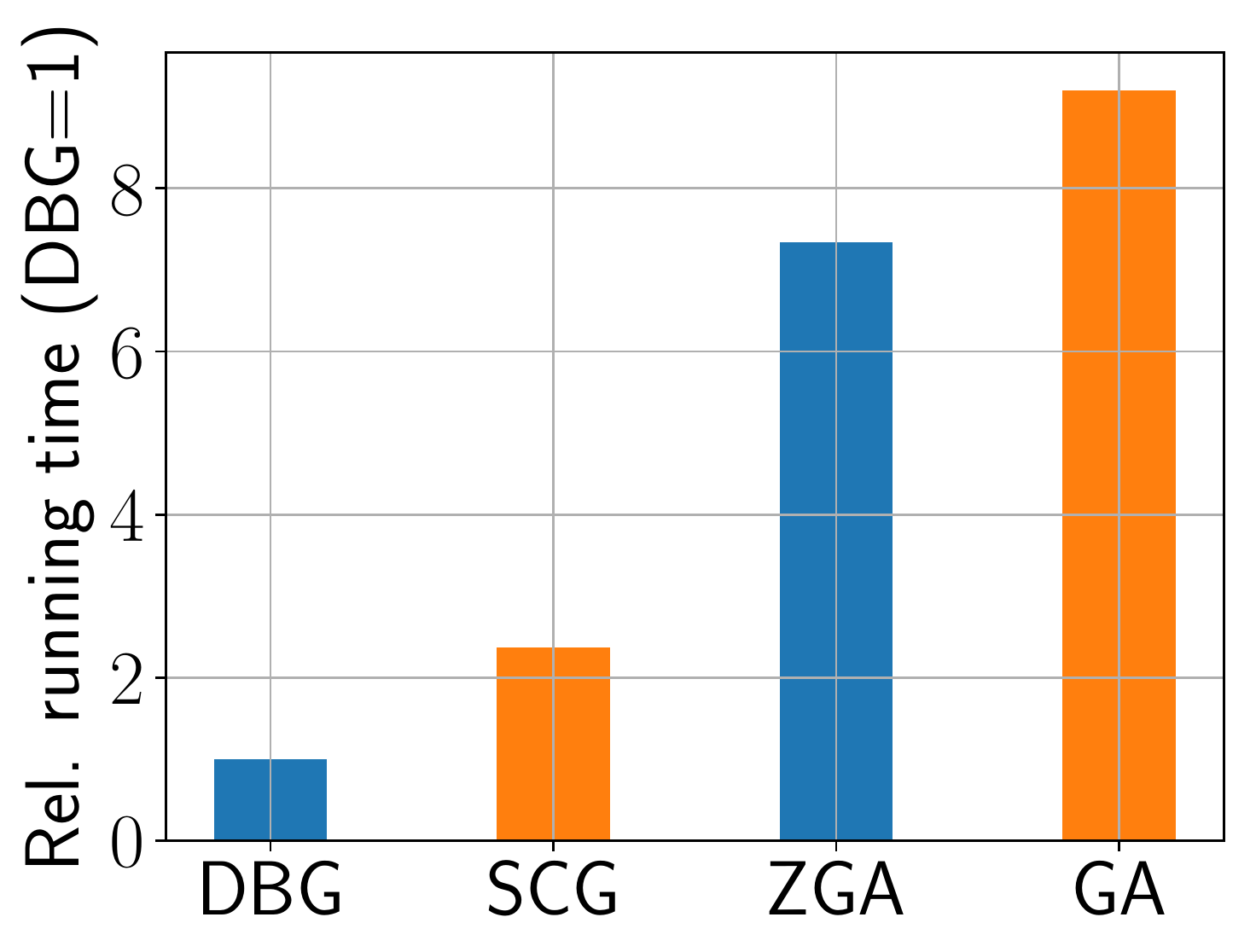}
		\caption{Influence maximization}
		\label{fig:cover_time}
	\end{subfigure}
	\caption{Relative running time normalized with respect to BCG (for 
		continuous 
		DR-submodular maximization in the first two sets of experiments) and 
		DBG 
		(for submodular set maximization in the last two sets of experiments). 
	}
\end{figure*}

We perform four sets of experiments which are described in detail in the following. The first two sets of 
experiments are maximization of continuous DR-submodular functions, which \Alg 
is designed to solve. The 
last two are submodular set maximization problems. We will apply \Algdis to solve 
these problems. 
The function values at 
different rounds and the execution times are presented in 
\cref{fig:function,fig:time}. The first-order algorithms (SCG and GA) are 
marked in \textcolor{orange}{orange}, and the zeroth-order algorithms are 
marked in \textcolor{RoyalBlue}{blue}.  


\textbf{Non-convex/non-concave Quadratic Programming (NQP):}
In this set of experiments, we apply our proposed algorithm and the baselines 
to the problem of non-convex/non-concave quadratic programming. The objective 
function is of the form $F(x) = \frac{1}{2}x^\top H x + b^\top x$, 
where $ x $ is a 100-dimensional vector, $ H $ is a $ 100$-by-$ 100 $ matrix, 
and every component of $H$ is an i.i.d.\ random variable whose distribution 
is equal to that of the negated absolute value of a standard normal 
distribution. 
The constraints are $ \sum_{i=1}^{30}x_i \le 30 $, $ \sum_{i=31}^{60} x_i \le 20 
$, and $ \sum_{i=61}^{100} x_i \le 20 $. To 
guarantee that the gradient is non-negative, we set $b_t=-H^\top \mathbf{1}$. 
 One can observe from \cref{fig:nqp_function} that 
the 
function 
value that BCG attains is only slightly lower than that of the first-order 
algorithm SCG. The final 
function value that BCG attains is similar to that of ZGA.

\textbf{Topic Summarization:} Next, we consider the topic 
summarization problem~\citep{el2009turning,yue2011linear}, which is to maximize 
the probabilistic coverage of selected articles on news topics. Each news 
article is characterized by its topic distribution, which is obtained by 
applying latent Dirichlet allocation to the corpus of Reuters-21578, 
Distribution 1.0. The number of topics is set to 10. We will choose from 120 
news articles. The probabilistic coverage of a subset of news articles (denoted 
by $ X $) is defined by
$f(X)=\frac{1}{10}\sum_{j=1}^{10}[1-\prod_{a \in X}(1-p_a(j))]$, where 
$p_a(\cdot)$ is the topic distribution of article 
$a$. 
The multilinear extension function of $f$ is 
$F(x)=\frac{1}{10}\sum_{j=1}^{10}[1-\prod_{a \in 
\Omega}(1-p_a(j)x_a)]$, where 
$x \in [0,1]^{120}$~\cite{Iyer2014Monotone}. The constraint is
$\sum_{i=1}^{40}x_i\leq 
25$, $\sum_{i=41}^{80}x_i\leq 30$, $\sum_{i=81}^{120}x_i\leq 35$. 
It can be observed from \cref{fig:topic_function} that the proposed BCG 
algorithm achieves the same function value as the first-ordered algorithm SCG 
and outperforms the other two. 
As shown in \cref{fig:nqp_time}, BCG is the most efficient method. The two 
projection-free algorithms BCG and SCG run faster than the projected methods 
ZGA and GA. We will elaborate on the running time later in this section.

\textbf{Active Set Selection}
We study the active set selection problem that arises in Gaussian process 
regression~\cite{mirzasoleiman2013distributed}. 
We use the \emph{Parkinsons
Telemonitoring} dataset, which is composed of biomedical voice measurements 
from people with early-stage Parkinson's disease \citep{tsanas2010enhanced}.
Let $ \mathbf{X}\in\mathbb{R}^{n\times d} $ denote the data matrix. Each row  $ 
X[i,:] $ is a voice recording while each column $ X[:,j] $ denotes an 
attribute. The covariance matrix $ \Sigma $ is 
defined by
 $ \Sigma_{ij} = \exp(-\| X[:,i]-X[:,j] \|^2)/h^2 $, where $ h $ is set to $ 
 0.75 $. The objective function of the active set selection problem is defined 
 by
 $ f(S) = \log\det(I+\Sigma_{S,S}) $, where $ S\subseteq [d] $ and $ 
 \Sigma_{S,S} $ is the principal submatrix indexed by $ S $. The total number 
 of 22 attributes are 
 partitioned into 5 disjoint subsets with sizes 4, 4, 4, 5 and 5, respectively. 
 The problem is 
 subject to a partition matroid requiring that at most one attribute should be 
 active within each subset.
 Since this is a submodular set maximization 
 problem, in order to evaluate the gradient (i.e., obtain an unbiased estimate 
 of gradient) required by first-order algorithms SCG and GA, it needs $ 2d $ 
 function value queries. To be precise, the $ i $-th component of gradient is $ 
 \expect_{S\sim x}[f(S\cup\{i\})-f(S)] $ and requires two function value 
 queries. It can be observed 
 from \cref{fig:active_function} that DBG outperforms the other zeroth-order 
 algorithm ZGA. Although its performance is slightly worse than the two 
 first-order algorithms SCG and GA, it require significantly less number of function value 
 queries than the other two first-order methods (as discussed above).
 
 \textbf{Influence Maximization}
In the influence maximization problem, we assume that every node in the network is able to influence all  of its one-hop 
neighbors.   
The objective of influence maximization is to select a subset of nodes in the network, called the seed set 
(and denoted by $ S $), so that the total number of  influenced nodes, 
including the seed nodes, is maximized. We choose the social network of 
Zachary's karate club~\cite{zachary1977information} in this study. The subjects 
in this social network are partitioned into three disjoint groups, whose sizes 
are 10, 14, and 10 respectively.  The chosen seed nodes should be subject to a 
partition matroid; i.e., 
 We will select at most two subjects from each of the 
three groups. Note that this problem is also a submodular set maximization 
problem. Similar to the situation in the active set selection problem, 
first-order algorithms need function value queries to obtain an unbiased 
estimate of gradient. We can observe from \cref{fig:cover_function} that DBG 
attains a better influence coverage than the other zeroth-order algorithm ZGA. 
Again, even though SCG and GA achieve a slightly better coverage, due to their 
first-order nature, they require a significantly larger number of function 
value queries.

\paragraph{Running Time}
The running times of the our proposed algorithms and the baselines   are
presented in \cref{fig:time} for the above-mentioned experimental set-ups. There are two main conclusions. First, the two 
projection-based algorithms (ZGA and GA) require significantly higher time complexity compared to the projection-free algorithms (BCG, DBG, and SCG), as the projection-based algorithms require 
solving quadratic optimization problems whereas projection-free ones 
require solving linear optimization problems which can be solved  more 
efficiently. Second, when we compare first-order and zeroth-order algorithms, we can 
observe that zeroth-order algorithms (BCG, DBG, and ZGA) run faster than their 
first-order counterparts (SCG and GA).
\paragraph{Summary}
\revise{The above experiment results show the following major 
advantages of our method over the baselines including SCG and ZGA.} 
\begin{itemize}
\item BCG/DBG is at least twice faster than SCG and ZGA in all 
tasks in terms of running time 
(\cref{fig:nqp_time,fig:topic_time,fig:active_time,fig:cover_time})
\item DBG requires remarkably fewer function evaluations in the discrete
setting (\cref{fig:active_function,fig:cover_function})
\item In addition to saving function evaluations, BCG/DBG achieves an objective 
function value comparable to that of the first-order baselines SCG and GA. 
\end{itemize}

\revise{Furthermore, we note that the number of first-order queries 
required by SCG 
is only half the number required by BCG. However, as is shown in 
\cref{fig:nqp_time,fig:topic_time}, BCG runs significantly faster than SCG 
since a zeroth-order evaluation is faster than a first-order one.} 

\revise{In the topic summarization task (\cref{fig:topic_function}), 
BCG exhibits a similar performance to that of the first-order baselines SCG and 
GA, in terms of the attained objective function value. In the other three 
tasks, BCG/DBG runs notably faster while achieving an only slightly inferior 
function value. 
Therefore, BCG/DBG is  particularly 
preferable  in a large-scale machine learning task and an application where the 
total number of function evaluations or 
the running time is subject to a budget. 
}

\section{Conclusion}
In this paper, we presented \Alg, a derivative-free and projection-free  
algorithm  for maximizing a 
monotone and  continuous 
DR-submodular function subject to a general convex body constraint.  We showed 
that \Alg achieves 
the tight 
$[(1-1/e)OPT-\epsilon]$ 
approximation guarantee with $\mathcal{O}(d/\epsilon^3)$ function evaluations. 
We then 
extended the algorithm to the stochastic continuous setting and the discrete 
submodular 
maximization 
problem. Our experiments on both synthetic and real data validated the 
performance of our proposed algorithms. In particular, we observed that \Alg 
practically achieves the 
same utility as \AlgCG while being way more efficient in terms of number of 
function evaluations.

\section*{Acknowledgements}
LC is supported by the Google PhD Fellowship. HH is supported by AFOSR Award 
19RT0726, NSF HDR TRIPODS award 1934876, NSF award CPS-1837253, NSF award 
CIF-1910056, and NSF CAREER award CIF-1943064. AK is partially 
supported by  NSF (IIS-1845032), ONR (N00014-19-1-2406), and AFOSR 
(FA9550-18-1-0160).

\bibliographystyle{plainnat}
\bibliography{reference-list}

\begin{thebibliography}{75}
\providecommand{\natexlab}[1]{#1}
\providecommand{\url}[1]{\texttt{#1}}
\expandafter\ifx\csname urlstyle\endcsname\relax
  \providecommand{\doi}[1]{doi: #1}\else
  \providecommand{\doi}{doi: \begingroup \urlstyle{rm}\Url}\fi

\bibitem[Agarwal et~al.(2010)Agarwal, Dekel, and Xiao]{agarwal2010optimal}
Alekh Agarwal, Ofer Dekel, and Lin Xiao.
\newblock Optimal algorithms for online convex optimization with multi-point
  bandit feedback.
\newblock In \emph{COLT}, pages 28--40. Citeseer, 2010.

\bibitem[Ageev and Sviridenko(2004)]{ageev2004pipage}
Alexander~A Ageev and Maxim~I Sviridenko.
\newblock Pipage rounding: A new method of constructing algorithms with proven
  performance guarantee.
\newblock \emph{Journal of Combinatorial Optimization}, 8\penalty0
  (3):\penalty0 307--328, 2004.

\bibitem[Audet and Hare(2017)]{audet2017derivative}
Charles Audet and Warren Hare.
\newblock \emph{Derivative-free and blackbox optimization}.
\newblock Springer, 2017.

\bibitem[Bach(2016)]{bach2016submodular}
Francis Bach.
\newblock Submodular functions: from discrete to continuous domains.
\newblock \emph{Mathematical Programming}, pages 1--41, 2016.

\bibitem[Bach et~al.(2012)Bach, Jenatton, Mairal, Obozinski,
  et~al.]{bach2012optimization}
Francis Bach, Rodolphe Jenatton, Julien Mairal, Guillaume Obozinski, et~al.
\newblock Optimization with sparsity-inducing penalties.
\newblock \emph{Foundations and Trends{\textregistered} in Machine Learning},
  4\penalty0 (1):\penalty0 1--106, 2012.

\bibitem[Bach(2010)]{bach2010structured}
Francis~R Bach.
\newblock Structured sparsity-inducing norms through submodular functions.
\newblock In \emph{Advances in Neural Information Processing Systems}, pages
  118--126, 2010.

\bibitem[Balasubramanian and Ghadimi(2018)]{balasubramanian2018zeroth}
Krishnakumar Balasubramanian and Saeed Ghadimi.
\newblock Zeroth-order (non)-convex stochastic optimization via conditional
  gradient and gradient updates.
\newblock In \emph{Advances in Neural Information Processing Systems}, pages
  3459--3468, 2018.

\bibitem[Balkanski and Singer(2015)]{balkanski2015mechanisms}
Eric Balkanski and Yaron Singer.
\newblock Mechanisms for fair attribution.
\newblock In \emph{Proceedings of the Sixteenth ACM Conference on Economics and
  Computation}, pages 529--546. ACM, 2015.

\bibitem[Bateni et~al.(2019)Bateni, Chen, Esfandiari, Fu, Mirrokni, and
  Rostamizadeh]{bateni2019categorical}
Mohammadhossein Bateni, Lin Chen, Hossein Esfandiari, Thomas Fu, Vahab
  Mirrokni, and Afshin Rostamizadeh.
\newblock Categorical feature compression via submodular optimization.
\newblock In \emph{International Conference on Machine Learning}, pages
  515--523, 2019.

\bibitem[Bergstra et~al.(2011)Bergstra, Bardenet, Bengio, and
  K{\'e}gl]{bergstra2011algorithms}
James~S Bergstra, R{\'e}mi Bardenet, Yoshua Bengio, and Bal{\'a}zs K{\'e}gl.
\newblock Algorithms for hyper-parameter optimization.
\newblock In \emph{Advances in neural information processing systems}, pages
  2546--2554, 2011.

\bibitem[Bian et~al.(2017{\natexlab{a}})Bian, Levy, Krause, and
  Buhmann]{bian2017continuous}
An~Bian, Kfir Levy, Andreas Krause, and Joachim~M Buhmann.
\newblock Continuous dr-submodular maximization: Structure and algorithms.
\newblock In \emph{Advances in Neural Information Processing Systems}, pages
  486--496, 2017{\natexlab{a}}.

\bibitem[Bian et~al.(2018)Bian, Buhmann, and Krause]{bian2018optimal}
An~Bian, Joachim~M Buhmann, and Andreas Krause.
\newblock Optimal dr-submodular maximization and applications to provable mean
  field inference.
\newblock \emph{arXiv preprint arXiv:1805.07482}, 2018.

\bibitem[Bian et~al.(2017{\natexlab{b}})Bian, Mirzasoleiman, Buhmann, and
  Krause]{bian2017guaranteed}
Andrew~An Bian, Baharan Mirzasoleiman, Joachim Buhmann, and Andreas Krause.
\newblock Guaranteed non-convex optimization: Submodular maximization over
  continuous domains.
\newblock In \emph{Artificial Intelligence and Statistics}, pages 111--120,
  2017{\natexlab{b}}.

\bibitem[Calinescu et~al.(2011)Calinescu, Chekuri, P{\'a}l, and
  Vondr{\'a}k]{calinescu2011maximizing}
Gruia Calinescu, Chandra Chekuri, Martin P{\'a}l, and Jan Vondr{\'a}k.
\newblock Maximizing a monotone submodular function subject to a matroid
  constraint.
\newblock \emph{SIAM Journal on Computing}, 40\penalty0 (6):\penalty0
  1740--1766, 2011.

\bibitem[Celis et~al.(2016)Celis, Deshpande, Kathuria, and
  Vishnoi]{celis2016fair}
L~Elisa Celis, Amit Deshpande, Tarun Kathuria, and Nisheeth~K Vishnoi.
\newblock How to be fair and diverse?
\newblock \emph{arXiv preprint arXiv:1610.07183}, 2016.

\bibitem[Chekuri et~al.(2014)Chekuri, Vondr{\'a}k, and
  Zenklusen]{chekuri2014submodular}
Chandra Chekuri, Jan Vondr{\'a}k, and Rico Zenklusen.
\newblock Submodular function maximization via the multilinear relaxation and
  contention resolution schemes.
\newblock \emph{SIAM Journal on Computing}, 43\penalty0 (6):\penalty0
  1831--1879, 2014.

\bibitem[Chen et~al.(2017{\natexlab{a}})Chen, Crawford, and
  Karbasi]{chen2017submodular}
Lin Chen, Forrest~W Crawford, and Amin Karbasi.
\newblock Submodular variational inference for network reconstruction.
\newblock In \emph{UAI}, 2017{\natexlab{a}}.

\bibitem[Chen et~al.(2017{\natexlab{b}})Chen, Krause, and
  Karbasi]{chen2017interactive}
Lin Chen, Andreas Krause, and Amin Karbasi.
\newblock Interactive submodular bandit.
\newblock In \emph{NeurIPS}, pages 141--152, 2017{\natexlab{b}}.

\bibitem[Chen et~al.(2018{\natexlab{a}})Chen, Feldman, and
  Karbasi]{chen2018weakly}
Lin Chen, Moran Feldman, and Amin Karbasi.
\newblock Weakly submodular maximization beyond cardinality constraints: Does
  randomization help greedy?
\newblock In \emph{ICML}, pages 804--813, 2018{\natexlab{a}}.

\bibitem[Chen et~al.(2018{\natexlab{b}})Chen, Harshaw, Hassani, and
  Karbasi]{pmlr-v80-chen18c}
Lin Chen, Christopher Harshaw, Hamed Hassani, and Amin Karbasi.
\newblock Projection-free online optimization with stochastic gradient: From
  convexity to submodularity.
\newblock In \emph{ICML}, pages 813--822, 10--15 Jul 2018{\natexlab{b}}.

\bibitem[Chen et~al.(2018{\natexlab{c}})Chen, Hassani, and
  Karbasi]{chen2018online}
Lin Chen, Hamed Hassani, and Amin Karbasi.
\newblock Online continuous submodular maximization.
\newblock In \emph{AISTATS}, pages 1896--1905, 2018{\natexlab{c}}.

\bibitem[Chen et~al.(2017{\natexlab{c}})Chen, Zhang, Sharma, Yi, and
  Hsieh]{chen2017zoo}
Pin-Yu Chen, Huan Zhang, Yash Sharma, Jinfeng Yi, and Cho-Jui Hsieh.
\newblock Zoo: Zeroth order optimization based black-box attacks to deep neural
  networks without training substitute models.
\newblock In \emph{Proceedings of the 10th ACM Workshop on Artificial
  Intelligence and Security}, pages 15--26. ACM, 2017{\natexlab{c}}.

\bibitem[Conn et~al.(2009)Conn, Scheinberg, and Vicente]{conn2009introduction}
Andrew~R Conn, Katya Scheinberg, and Luis~N Vicente.
\newblock \emph{Introduction to derivative-free optimization}, volume~8.
\newblock Siam, 2009.

\bibitem[Das and Kempe(2011)]{das2011submodular}
Abhimanyu Das and David Kempe.
\newblock Submodular meets spectral: Greedy algorithms for subset selection,
  sparse approximation and dictionary selection.
\newblock \emph{arXiv preprint arXiv:1102.3975}, 2011.

\bibitem[Eghbali and Fazel(2016)]{eghbali2016designing}
Reza Eghbali and Maryam Fazel.
\newblock Designing smoothing functions for improved worst-case competitive
  ratio in online optimization.
\newblock In \emph{Advances in Neural Information Processing Systems}, pages
  3287--3295, 2016.

\bibitem[El-Arini et~al.(2009)El-Arini, Veda, Shahaf, and
  Guestrin]{el2009turning}
Khalid El-Arini, Gaurav Veda, Dafna Shahaf, and Carlos Guestrin.
\newblock Turning down the noise in the blogosphere.
\newblock In \emph{Proceedings of the 15th ACM SIGKDD international conference
  on Knowledge discovery and data mining}, pages 289--298. ACM, 2009.

\bibitem[Flaxman et~al.(2005)Flaxman, Kalai, and McMahan]{flaxman2005online}
Abraham~D Flaxman, Adam~Tauman Kalai, and H~Brendan McMahan.
\newblock Online convex optimization in the bandit setting: gradient descent
  without a gradient.
\newblock In \emph{Proceedings of the sixteenth annual ACM-SIAM symposium on
  Discrete algorithms}, pages 385--394. Society for Industrial and Applied
  Mathematics, 2005.

\bibitem[Frank and Wolfe(1956)]{frank1956algorithm}
Marguerite Frank and Philip Wolfe.
\newblock An algorithm for quadratic programming.
\newblock \emph{Naval research logistics quarterly}, 3\penalty0 (1-2):\penalty0
  95--110, 1956.

\bibitem[Ghadimi and Lan(2013)]{ghadimi2013stochastic}
Saeed Ghadimi and Guanghui Lan.
\newblock Stochastic first-and zeroth-order methods for nonconvex stochastic
  programming.
\newblock \emph{SIAM Journal on Optimization}, 23\penalty0 (4):\penalty0
  2341--2368, 2013.

\bibitem[Golovin and Krause(2011)]{golovin2011adaptive}
Daniel Golovin and Andreas Krause.
\newblock Adaptive submodularity: Theory and applications in active learning
  and stochastic optimization.
\newblock \emph{Journal of Artificial Intelligence Research}, 42:\penalty0
  427--486, 2011.

\bibitem[Guillory and Bilmes(2010)]{guillory2010interactive}
Andrew Guillory and Jeff Bilmes.
\newblock Interactive submodular set cover.
\newblock \emph{arXiv preprint arXiv:1002.3345}, 2010.

\bibitem[Hassani et~al.(2017)Hassani, Soltanolkotabi, and
  Karbasi]{hassani2017gradient}
Hamed Hassani, Mahdi Soltanolkotabi, and Amin Karbasi.
\newblock Gradient methods for submodular maximization.
\newblock In \emph{Advances in Neural Information Processing Systems}, pages
  5841--5851, 2017.

\bibitem[Hassani et~al.(2019)Hassani, Karbasi, Mokhtari, and
  Shen]{hassani2019stochastic}
Hamed Hassani, Amin Karbasi, Aryan Mokhtari, and Zebang Shen.
\newblock Stochastic continuous greedy++: When upper and lower bounds match.
\newblock In \emph{Advances in Neural Information Processing Systems}, pages
  13066--13076, 2019.

\bibitem[Hazan(2016)]{hazan2016introduction}
Elad Hazan.
\newblock Introduction to online convex optimization.
\newblock \emph{Foundations and Trends{\textregistered} in Optimization},
  2\penalty0 (3-4):\penalty0 157--325, 2016.

\bibitem[Hazan and Luo(2016)]{hazan2016variance}
Elad Hazan and Haipeng Luo.
\newblock Variance-reduced and projection-free stochastic optimization.
\newblock In \emph{International Conference on Machine Learning}, pages
  1263--1271, 2016.

\bibitem[Ilyas et~al.(2018)Ilyas, Engstrom, Athalye, Lin, Athalye, Engstrom,
  Ilyas, and Kwok]{ilyas2018black}
Andrew Ilyas, Logan Engstrom, Anish Athalye, Jessy Lin, Anish Athalye, Logan
  Engstrom, Andrew Ilyas, and Kevin Kwok.
\newblock Black-box adversarial attacks with limited queries and information.
\newblock In \emph{Proceedings of the 35th International Conference on Machine
  Learning,$\{$ICML$\}$ 2018}, 2018.

\bibitem[Iyer et~al.(2014)Iyer, Jegelka, and Bilmes]{Iyer2014Monotone}
Rishabh Iyer, Stefanie Jegelka, and Jeff Bilmes.
\newblock Monotone closure of relaxed constraints in submodular optimization:
  Connections between minimization and maximization.
\newblock In \emph{Uncertainty in Artificial Intelligence (UAI)}, Quebic City,
  Quebec Canada, July 2014. AUAI.

\bibitem[Jaggi(2013)]{jaggi2013revisiting}
Martin Jaggi.
\newblock Revisiting frank-wolfe: Projection-free sparse convex optimization.
\newblock In \emph{ICML (1)}, pages 427--435, 2013.

\bibitem[Jegelka and Bilmes(2011{\natexlab{a}})]{jegelka2011submodularity}
Stefanie Jegelka and Jeff Bilmes.
\newblock Submodularity beyond submodular energies: coupling edges in graph
  cuts.
\newblock 2011{\natexlab{a}}.

\bibitem[Jegelka and Bilmes(2011{\natexlab{b}})]{jegelka2011approximation}
Stefanie Jegelka and Jeff~A Bilmes.
\newblock Approximation bounds for inference using cooperative cuts.
\newblock In \emph{Proceedings of the 28th International Conference on Machine
  Learning (ICML-11)}, pages 577--584, 2011{\natexlab{b}}.

\bibitem[Kempe et~al.(2003)Kempe, Kleinberg, and Tardos]{kempe2003maximizing}
David Kempe, Jon Kleinberg, and {\'E}va Tardos.
\newblock Maximizing the spread of influence through a social network.
\newblock In \emph{Proceedings of the ninth ACM SIGKDD international conference
  on Knowledge discovery and data mining}, pages 137--146. ACM, 2003.

\bibitem[Kulesza et~al.(2012)Kulesza, Taskar, et~al.]{kulesza2012determinantal}
Alex Kulesza, Ben Taskar, et~al.
\newblock Determinantal point processes for machine learning.
\newblock \emph{Foundations and Trends{\textregistered} in Machine Learning},
  5\penalty0 (2--3):\penalty0 123--286, 2012.

\bibitem[Lacoste-Julien and Jaggi(2015)]{lacoste2015global}
Simon Lacoste-Julien and Martin Jaggi.
\newblock On the global linear convergence of frank-wolfe optimization
  variants.
\newblock In \emph{Advances in Neural Information Processing Systems}, pages
  496--504, 2015.

\bibitem[Lei et~al.(2019)Lei, Wu, Chen, Dimakis, Dhillon, and
  Witbrock]{lei2019discrete}
Qi~Lei, Lingfei Wu, Pin-Yu Chen, Alexandros Dimakis, Inderjit Dhillon, and
  Michael Witbrock.
\newblock Discrete adversarial attacks and submodular optimization with
  applications to text classification.
\newblock \emph{Systems and Machine Learning (SysML)}, 2019.

\bibitem[Lin and Bilmes(2011{\natexlab{a}})]{lin2011class}
Hui Lin and Jeff Bilmes.
\newblock A class of submodular functions for document summarization.
\newblock In \emph{Proceedings of the 49th Annual Meeting of the Association
  for Computational Linguistics: Human Language Technologies-Volume 1}, pages
  510--520. Association for Computational Linguistics, 2011{\natexlab{a}}.

\bibitem[Lin and Bilmes(2011{\natexlab{b}})]{lin2011word}
Hui Lin and Jeff Bilmes.
\newblock Word alignment via submodular maximization over matroids.
\newblock In \emph{Proceedings of the 49th Annual Meeting of the Association
  for Computational Linguistics: Human Language Technologies: short
  papers-Volume 2}, pages 170--175. Association for Computational Linguistics,
  2011{\natexlab{b}}.

\bibitem[Mirzasoleiman et~al.(2013)Mirzasoleiman, Karbasi, Sarkar, and
  Krause]{mirzasoleiman2013distributed}
Baharan Mirzasoleiman, Amin Karbasi, Rik Sarkar, and Andreas Krause.
\newblock Distributed submodular maximization: Identifying representative
  elements in massive data.
\newblock In \emph{Advances in Neural Information Processing Systems}, pages
  2049--2057, 2013.

\bibitem[Mokhtari et~al.(2018{\natexlab{a}})Mokhtari, Hassani, and
  Karbasi]{mokhtari2018conditional}
Aryan Mokhtari, Hamed Hassani, and Amin Karbasi.
\newblock Conditional gradient method for stochastic submodular maximization:
  Closing the gap.
\newblock In \emph{AISTATS}, pages 1886--1895, 2018{\natexlab{a}}.

\bibitem[Mokhtari et~al.(2018{\natexlab{b}})Mokhtari, Hassani, and
  Karbasi]{mokhtari2018stochastic}
Aryan Mokhtari, Hamed Hassani, and Amin Karbasi.
\newblock Stochastic conditional gradient methods: From convex minimization to
  submodular maximization.
\newblock \emph{arXiv preprint arXiv:1804.09554}, 2018{\natexlab{b}}.

\bibitem[Nemhauser et~al.(1978)Nemhauser, Wolsey, and
  Fisher]{nemhauser1978analysis}
George~L Nemhauser, Laurence~A Wolsey, and Marshall~L Fisher.
\newblock An analysis of approximations for maximizing submodular set
  functions---i.
\newblock \emph{Mathematical programming}, 14\penalty0 (1):\penalty0 265--294,
  1978.

\bibitem[Rios and Sahinidis(2013)]{rios2013derivative}
Luis~Miguel Rios and Nikolaos~V Sahinidis.
\newblock Derivative-free optimization: a review of algorithms and comparison
  of software implementations.
\newblock \emph{Journal of Global Optimization}, 56\penalty0 (3):\penalty0
  1247--1293, 2013.

\bibitem[Rodriguez and Sch{\"o}lkopf(2012)]{rodriguez2012influence}
Manuel~Gomez Rodriguez and Bernhard Sch{\"o}lkopf.
\newblock Influence maximization in continuous time diffusion networks.
\newblock \emph{arXiv preprint arXiv:1205.1682}, 2012.

\bibitem[Sahu et~al.(2018)Sahu, Zaheer, and Kar]{sahu2018towards}
Anit~Kumar Sahu, Manzil Zaheer, and Soummya Kar.
\newblock Towards gradient free and projection free stochastic optimization.
\newblock \emph{arXiv preprint arXiv:1810.03233}, 2018.

\bibitem[Salehi et~al.(2017)Salehi, Karbasi, Scheinost, and
  Constable]{salehi2017submodular}
Mehraveh Salehi, Amin Karbasi, Dustin Scheinost, and R~Todd Constable.
\newblock A submodular approach to create individualized parcellations of the
  human brain.
\newblock In \emph{International Conference on Medical Image Computing and
  Computer-Assisted Intervention}, pages 478--485. Springer, 2017.

\bibitem[Shahriari et~al.(2016)Shahriari, Swersky, Wang, Adams, and
  De~Freitas]{shahriari2016taking}
Bobak Shahriari, Kevin Swersky, Ziyu Wang, Ryan~P Adams, and Nando De~Freitas.
\newblock Taking the human out of the loop: A review of bayesian optimization.
\newblock \emph{Proceedings of the IEEE}, 104\penalty0 (1):\penalty0 148--175,
  2016.

\bibitem[Shamir(2017)]{shamir2017optimal}
Ohad Shamir.
\newblock An optimal algorithm for bandit and zero-order convex optimization
  with two-point feedback.
\newblock \emph{Journal of Machine Learning Research}, 18\penalty0
  (52):\penalty0 1--11, 2017.

\bibitem[Singla et~al.(2014)Singla, Bogunovic, Bart{\'o}k, Karbasi, and
  Krause]{singla2014near}
Adish Singla, Ilija Bogunovic, G{\'a}bor Bart{\'o}k, Amin Karbasi, and Andreas
  Krause.
\newblock Near-optimally teaching the crowd to classify.
\newblock In \emph{ICML}, pages 154--162, 2014.

\bibitem[Snoek et~al.(2012)Snoek, Larochelle, and Adams]{snoek2012practical}
Jasper Snoek, Hugo Larochelle, and Ryan~P Adams.
\newblock Practical bayesian optimization of machine learning algorithms.
\newblock In \emph{Advances in neural information processing systems}, pages
  2951--2959, 2012.

\bibitem[Sokolov et~al.(2016)Sokolov, Kreutzer, Riezler, and
  Lo]{sokolov2016stochastic}
Artem Sokolov, Julia Kreutzer, Stefan Riezler, and Christopher Lo.
\newblock Stochastic structured prediction under bandit feedback.
\newblock In \emph{Advances in Neural Information Processing Systems}, pages
  1489--1497, 2016.

\bibitem[Staib and Jegelka(2017)]{staib2017robust}
Matthew Staib and Stefanie Jegelka.
\newblock Robust budget allocation via continuous submodular functions.
\newblock \emph{arXiv preprint arXiv:1702.08791}, 2017.

\bibitem[Steudel et~al.(2010)Steudel, Janzing, and
  Sch{\"o}lkopf]{steudel2010causal}
Bastian Steudel, Dominik Janzing, and Bernhard Sch{\"o}lkopf.
\newblock Causal markov condition for submodular information measures.
\newblock \emph{arXiv preprint arXiv:1002.4020}, 2010.

\bibitem[Taskar et~al.(2005)Taskar, Chatalbashev, Koller, and
  Guestrin]{taskar2005learning}
Ben Taskar, Vassil Chatalbashev, Daphne Koller, and Carlos Guestrin.
\newblock Learning structured prediction models: A large margin approach.
\newblock In \emph{Proceedings of the 22nd international conference on Machine
  learning}, pages 896--903. ACM, 2005.

\bibitem[Thornton et~al.(2013)Thornton, Hutter, Hoos, and
  Leyton-Brown]{thornton2013auto}
Chris Thornton, Frank Hutter, Holger~H Hoos, and Kevin Leyton-Brown.
\newblock Auto-weka: Combined selection and hyperparameter optimization of
  classification algorithms.
\newblock In \emph{Proceedings of the 19th ACM SIGKDD international conference
  on Knowledge discovery and data mining}, pages 847--855. ACM, 2013.

\bibitem[Tsanas et~al.(2010)Tsanas, Little, McSharry, and
  Ramig]{tsanas2010enhanced}
Athanasios Tsanas, Max~A Little, Patrick~E McSharry, and Lorraine~O Ramig.
\newblock Enhanced classical dysphonia measures and sparse regression for
  telemonitoring of parkinson's disease progression.
\newblock In \emph{Acoustics Speech and Signal Processing (ICASSP), 2010 IEEE
  International Conference on}, pages 594--597. IEEE, 2010.

\bibitem[Tschiatschek et~al.(2014)Tschiatschek, Iyer, Wei, and
  Bilmes]{tschiatschek2014learning}
Sebastian Tschiatschek, Rishabh~K Iyer, Haochen Wei, and Jeff~A Bilmes.
\newblock Learning mixtures of submodular functions for image collection
  summarization.
\newblock In \emph{Advances in neural information processing systems}, pages
  1413--1421, 2014.

\bibitem[Vondr{\'a}k(2008)]{vondrak2008optimal}
Jan Vondr{\'a}k.
\newblock Optimal approximation for the submodular welfare problem in the value
  oracle model.
\newblock In \emph{Proceedings of the fortieth annual ACM symposium on Theory
  of computing}, pages 67--74. ACM, 2008.

\bibitem[Wainwright et~al.(2008)Wainwright, Jordan,
  et~al.]{wainwright2008graphical}
Martin~J Wainwright, Michael~I Jordan, et~al.
\newblock Graphical models, exponential families, and variational inference.
\newblock \emph{Foundations and Trends{\textregistered} in Machine Learning},
  1\penalty0 (1--2):\penalty0 1--305, 2008.

\bibitem[Wang et~al.(2017)Wang, Du, Balakrishnan, and
  Singh]{wang2017stochastic}
Yining Wang, Simon Du, Sivaraman Balakrishnan, and Aarti Singh.
\newblock Stochastic zeroth-order optimization in high dimensions.
\newblock \emph{arXiv preprint arXiv:1710.10551}, 2017.

\bibitem[Wei et~al.(2015)Wei, Iyer, and Bilmes]{wei2015submodularity}
Kai Wei, Rishabh Iyer, and Jeff Bilmes.
\newblock Submodularity in data subset selection and active learning.
\newblock In \emph{International Conference on Machine Learning}, pages
  1954--1963, 2015.

\bibitem[Yue and Guestrin(2011)]{yue2011linear}
Yisong Yue and Carlos Guestrin.
\newblock Linear submodular bandits and their application to diversified
  retrieval.
\newblock In \emph{Advances in Neural Information Processing Systems}, pages
  2483--2491, 2011.

\bibitem[Zachary(1977)]{zachary1977information}
Wayne~W Zachary.
\newblock An information flow model for conflict and fission in small groups.
\newblock \emph{Journal of anthropological research}, 33\penalty0 (4):\penalty0
  452--473, 1977.

\bibitem[Zhang et~al.(2019{\natexlab{a}})Zhang, Chen, Hassani, and
  Karbasi]{zhang2019online}
Mingrui Zhang, Lin Chen, Hamed Hassani, and Amin Karbasi.
\newblock Online continuous submodular maximization: From full-information to
  bandit feedback.
\newblock In \emph{Advances in Neural Information Processing Systems}, pages
  9206--9217, 2019{\natexlab{a}}.

\bibitem[Zhang et~al.(2019{\natexlab{b}})Zhang, Shen, Mokhtari, Hassani, and
  Karbasi]{zhang2019one}
Mingrui Zhang, Zebang Shen, Aryan Mokhtari, Hamed Hassani, and Amin Karbasi.
\newblock One sample stochastic frank-wolfe.
\newblock \emph{arXiv preprint arXiv:1910.04322}, 2019{\natexlab{b}}.

\bibitem[Zhang et~al.(2016)Zhang, Bai, Chen, Bian, and Li]{zhang2016influence}
Yuanxing Zhang, Yichong Bai, Lin Chen, Kaigui Bian, and Xiaoming Li.
\newblock Influence maximization in messenger-based social networks.
\newblock In \emph{GLOBECOM}, pages 1--6. IEEE, 2016.

\bibitem[Zhou and Spanos(2016)]{zhou2016causal}
Yuxun Zhou and Costas~J Spanos.
\newblock Causal meets submodular: Subset selection with directed information.
\newblock In \emph{Advances in Neural Information Processing Systems}, pages
  2649--2657, 2016.

\end{thebibliography}

\clearpage 
\onecolumn
\appendix
\section{Proof of \cref{lem:smooth_approx}}\label{app:proof_lemma_smooth_approx}
\begin{proof}
	Using %
	the assumption that
	$F$ is $G$-Lipschitz continuous, we have
	\begin{align}
	|\tF(x)-\tF(y)| =& |\expect_{v\sim B^d} [F(x+\delta v) - F(y+\delta v)]| \\
	\le& \expect_{v\sim B^d}[ |F(x+\delta v) - F(y+\delta v)|] \\
	\le & \expect_{v\sim B^d}[G\|(x+\delta v)-(y+\delta v)\|] \\
	=& G\|x-y\|,
	\end{align}
	and
	\begin{align}
	|\tilde{F}(x)-F(x)| = & | \expect_{v\sim B^d}[F(x+\delta v)-F(x)] | \\
	\leq & \expect_{v\sim B^d} [|F(x+\delta v)-F(x)|] \\
	\leq & \expect_{v\sim B^d}[G \delta \|v \|] \\
	\le & \delta G.
	\end{align}
	
	If $F$ is $G$-Lipschitz continuous and monotone continuous DR-submodular, 
	then 
	$F$ is differentiable. For $\forall x \leq y$, we also have
	\begin{equation}
	\nabla F(x) \geq \nabla F(y),   
	\end{equation}
	and
	\begin{equation}
	F(x) \leq F(y).    
	\end{equation}
	
	By definition of $\tilde{F}$, we have $\tilde{F}$ is differentiable and for 
	$\forall x \leq y$,
	\begin{align}
	\nabla \tilde{F}(x) - \nabla \tilde{F}(y) = & \nabla \expect_{v\sim B^d} 
	[F(x+\delta v)]-  \nabla \expect_{v\sim B^d} 
	[F(y+\delta v)] \\
	= & \expect_{v\sim B^d}[\nabla F(x+\delta v) - \nabla F(y+\delta v) ] \\
	\geq & \expect_{v\sim B^d} [0] \\
	=& 0,
	\end{align}
	and
	\begin{align}
	\tF(x)-\tF(y) =& \expect_{v \sim B^d}[F(x+\delta v)]-\expect_{v \sim 
		B^d}[F(y+\delta v)] 
	\\
	=& \expect_{v \sim B^d}[F(x+\delta v)-F(y+\delta v)] \\
	\leq & \expect_{v \sim B^d}[0] \\
	=& 0,
	\end{align}
	\emph{i.e.}, $\nabla \tilde{F}(x) \geq \nabla \tilde{F}(y), \tF(x) \leq 
	\tF(y).$ So $\tF$ is also a monotone continuous DR-submodular function.
\end{proof}

\section{Proof of \cref{thm:zero}}\label{app:theorem_zero}
In order to prove \cref{thm:zero}, we need the following variance reduction 
lemmas~\citep{shamir2017optimal,pmlr-v80-chen18c}, where the second one is a 
slight improvement of Lemma~2 in~\citep{mokhtari2018conditional} and Lemma~5 
in~\citep{mokhtari2018stochastic}. 

\begin{lemma}[Lemma~10 of \citep{shamir2017optimal}]\label{lem:variance}
	It holds that
	\begin{equation}
	\expect_{u \sim S^{d-1}}[\frac{d}{2\delta}(F(z+\delta u)-F(z-\delta u))u| 
	z]  = \nabla \tilde{F}(z) ,  
	\end{equation}
	\begin{equation}
	\expect_{u \sim S^{d-1}}[\|\frac{d}{2\delta}(F(z+\delta u)-F(z-\delta 
	u))u - \nabla \tF(z) \|^2|z]   \le cdG^2,       
	\end{equation}
	where $c$ is a constant.
\end{lemma}

\begin{lemma}[Theorem~3 of 
	\citep{pmlr-v80-chen18c}]\label{lem:variance_reduction}
	Let $ \{ 
	a_t\}_{t=0}^{T}$ be a sequence of points in $\mathbb{R}^n$ 
	such that $ \| a_t - 
	a_{t-1} \| 
	\leq G_0/(t+s)  $ for all $1\leq t\leq T $ with fixed constants $ G_0 \geq 
	0 
	$ and $ s\geq 3 $. 
	Let $ \{ \tilde{a}_t\}_{t=1}^T$ be a sequence of random variables such that 
	$ \expect[ 
	\tilde{a}_t|\mathcal{F}_{t-1} ] = a_t $ and $ \expect[ \| \tilde{a}_t - a_t 
	\|^2|\mathcal{F}_{t-1} ] \leq \sigma^2$ for 
	every $ t\geq 0 $, 
	where 
	$\mathcal{F}_{t-1} $ is the $ \sigma $-field generated by 
	$ \{ \tilde{a}_i\}_{i=1}^{t} $ 
	and $ \mathcal{F}_{0} = \varnothing $. Let $\{d_t\}_{t=0}^T$ be a sequence 
	of random 
	variables where $d_0$ is fixed and subsequent $d_{t}$ are obtained by the 
	recurrence 
	\begin{equation}
	d_t = (1-\rho_t) d_{t-1} +\rho_t \tilde{a}_t
	\end{equation}
	with $ \rho_t = \frac{2}{(t+s)^{2/3}} $. 
	Then, we have
	\begin{equation}
	\expect[\| a_t-d_t\|^2 ] \leq \frac{Q}{(t+s+1)^{2/3}},
	\end{equation}
	where $ Q \triangleq \max \{ \|a_0 - d_0 \|^2 (s+1)^{2/3}, 
	4\sigma^2 + 3G_0^2/2 \} $.
\end{lemma}

Now we turn to prove \cref{thm:zero}.

\begin{proof}[Proof of \cref{thm:zero}]
	First of all, we note that technically we need the iteration number $T \geq 
	4$, which always holds in practical applications.
	
	Then we show that $\forall t=1,\dots, T+1$, $x_t\in \domainsh$. By 
	the definition of $x_t$, we have $x_t=\sum_{i=1}^{t-1}\frac{v_i}{T}$. Since 
	$v_t$'s are non-negative vectors, we know that $x_t$'s are also 
	non-negative vectors and that $0=x_1\le x_2 \le \dots \le x_{T+1}$. It 
	suffices to show that $x_{T+1}\in \domainsh$. Since $x_{T+1}$ is a convex 
	combination of $v_1,\dots, v_T$ and $v_t$'s are in $\domainsh$, we conclude 
	that $x_{T+1}\in \domainsh$. In addition, since $v_t$'s are also in 
	$\constraint-\delta \one$, $x_{T+1}$ is also in $\constraint-\delta\one$. 
	Therefore our final choice $x_{T+1}+\delta \one$ resides in the constraint 
	$\constraint$.
	
	Let $z_t\triangleq x_t+\delta \one$ and the shrunk domain (without 
	translation) 
	$\domainsh'\triangleq \domainsh + \delta \one = \prod_{i=1}^d [\delta, 
	a_i-\delta]\subseteq \domain$. %
	By Jensen's inequality and the fact $F$ has $L$-Lipschitz continuous 
	gradients, we have
	\begin{equation}
	\| \nabla \tF(x) -\nabla \tF(y) \| \le L\|x-y\|.    
	\end{equation}
	Thus,
	\begin{align}
	\tF(z_{t+1})-\tF(z_t)= & \tF(z_t+\frac{v_t}{T})-\tF(z_t)\\
	\ge & \frac{1}{T} \nabla \tF(z_t)^\top v_t - \frac{L}{2 T^2} \| v_t \|^2 \\
	\ge & \frac{1}{T} \nabla \tF(z_t)^\top v_t - \frac{L}{2 T^2} D_1^2 \\
	= & \frac{1}{T}\left(\bar{g}_t^\top v_t + (\nabla \tF(z_t)-\bar{g}_t)^\top 
	v_t \right)  - \frac{L}{2 T^2} D_1^2.\label{eq:gv}
	\end{align}
	Let $x^*_\delta \triangleq \argmax_{x\in \domainsh'\cap \constraint} 
	\tF(x)$.
	Since $x^*_\delta,z_t\in \domainsh'$, we have $v_t^*\triangleq (x^*_\delta 
	- z_t)\vee 0 \in \domainsh$. We know  $z_t+v_t^* = x_\delta^*\vee z_t\in 
	\domainsh'$ and 
	\begin{equation}
	v_t^*+\delta \one =(x_\delta^*-x_t)\vee \delta \one \le x_\delta^*.
	\end{equation}
	
	Since we assume that $F$ is monotone continuous DR-submodular, by 
	\cref{lem:smooth_approx}, $\tF$ is also monotone continuous DR-submodular. 
	As a result, $\tF$ is concave along non-negative directions, and $\nabla 
	\tF$ is 
	entry-wise non-negative. Thus we have
	\begin{align}
	\tF(z_t+v_t^*)-\tF(z_t)\le & \nabla \tF(z_t)^\top v_t^*\\
	\le & \nabla \tF(z_t)^\top (x_\delta^*-\delta\one).
	\end{align}
	Since $x_\delta^*-\delta\one\in \constraint'$, we deduce \begin{align}
	\bar{g}_t^\top v_t \ge & \bar{g}_t^\top (x_\delta^*-\delta\one)\\
	= & \nabla \tF(z_t)^\top (x_\delta^*-\delta\one) + (\bar{g}_t-\nabla 
	\tF(z_t))^\top (x_\delta^*-\delta\one)\\
	\ge & \tF(z_t+v_t^*)-\tF(z_t) + (\bar{g}_t-\nabla \tF(z_t))^\top 
	(x_\delta^*-\delta\one)\\
	\ge & \tF(x_\delta^*)-\tF(z_t) + (\bar{g}_t-\nabla \tF(z_t))^\top 
	(x_\delta^*-\delta\one).
	\end{align}
	Therefore, we obtain
	\begin{align}
	\bar{g}_t^\top v_t + (\nabla \tF(z_t)-\bar{g}_t)^\top v_t \ge  
	\tF(x_\delta^*)-\tF(z_t) + (\nabla \tF(z_t)-\bar{g}_t)^\top 
	(v_t-(x_\delta^*-\delta\one)).\label{eq:gv2}
	\end{align}
	By plugging \cref{eq:gv2} into \cref{eq:gv}, after re-arrangement of the 
	terms, we obtain
	\begin{align}\label{eq:h_t}
	h_{t+1} \le (1-\frac{1}{T})h_t + \frac{1}{T}(\nabla 
	\tF(z_t)-\bar{g}_t)^\top ((x_\delta^*-\delta\one)-v_t)
	+\frac{L}{2 T^2} D_1^2,
	\end{align}
	where $h_t\triangleq \tF(x_\delta^*)-\tF(z_t)$. Next we derive an upper 
	bound for $(\nabla \tF(z_t)-\bar{g}_t)^\top ((x_\delta^*-\delta\one)-v_t)$. 
	By Young's inequality, it can be deduced that for any $\beta_t>0$,
	\begin{align}
	(\nabla \tF(z_t)-\bar{g}_t)^\top ((x_\delta^*-\delta\one)-v_t) \le & 
	\frac{\beta_t}{2}\| \nabla \tF(z_t)-\bar{g}_t\|^2 + 
	\frac{1}{2\beta_t} \|(x_\delta^*-\delta\one)-v_t \|^2 \\
	\le & \frac{\beta_t}{2}\| \nabla \tF(z_t)-\bar{g}_t\|^2 + 
	\frac{1}{2\beta_t} D_1^2. \label{eq:inner_product}
	\end{align}

	Now let $\mathcal{F}_1 \triangleq \varnothing$ and $\mathcal{F}_t$ be the 
	$\sigma$-field generate by $\{\bar{g}_1,
	\dots,\bar{g}_{t-1} \}$, then by \cref{lem:variance}, we have
	\begin{equation}
	\expect[\frac{d}{2\delta}(F(y_{t,i}^+)-F(y_{t,i}^-))u_{t,i}|\mathcal{F}_{t-1}]
	= \nabla \tF(z_t) ,  
	\end{equation}
	and
	\begin{equation}
	\expect[\|\frac{d}{2\delta}(F(y_{t,i}^+)-F(y_{t,i}^-))u_{t,i}-\nabla 
	\tF(z_t) \|^2|\mathcal{F}_{t-1}]  \leq cdG^2.  
	\end{equation}
	
	Therefore,
	\begin{align}
	\expect[g_t|\mathcal{F}_{t-1}] 
	=&\expect[\frac{1}{B_t}\sum_{i=1}^{B_t}\frac{d}{2\delta}(F(y_{t,i}^+)-F(y_{t,i}^-))u_{t,i}|\mathcal{F}_{t-1}]
	\\
	=&\nabla \tF(z_t) , 
	\end{align}
	and
	\begin{align}
	\expect[\|g_t - \nabla \tF(z_t) \|^2|\mathcal{F}_{t-1}] =&\frac{1}{B_t^2} 
	\sum_{i=1}^{B_t} 
	\expect[\|\frac{d}{2\delta}(F(y_{t,i}^+)-F(y_{t,i}^-))u_{t,i}-\nabla 
	\tF(z_t) \|^2|\mathcal{F}_{t-1}] \\
	\leq& \frac{cdG^2}{B_t}.  \label{eq:variance_upper}   
	\end{align}
	By Jensen's inequality and the assumption $F$ is $L$-smooth, we have
	\begin{equation}
		\|\nabla \tF(z_t)-\nabla\tF(z_{t-1}) \|\le L\frac{D_1}{T}\le 
		\frac{2LD_1}{t+3}.
	\end{equation}
	Then by \cref{lem:variance_reduction} with $s=3, d_t = \bar{g}_t, \forall t 
	\ge 
	0, \tilde{a}_t=g_t, a_t=\nabla \tF(z_t), \forall t \ge 1, a_0=\nabla 
	\tF(z_1), G_0=2LD_1$, we have
	\begin{equation}
	\expect[\| \nabla \tF(z_t)-\bar{g}_t\|^2]\le 
	\frac{Q}{(t+4)^{2/3}},\label{eq:bound_on_norm}
	\end{equation}
	where $Q\triangleq \max \{ \|\nabla \tF(x_1+\delta \one) \|^2 4^{2/3}, 
	\frac{4cdG^2}{B_t}+ 6L^2D_1^2 \} $.
	Note that by \cref{lem:smooth_approx}, we have $\| \nabla \tF(x) \| \le G$, 
	thus we can re-define $Q = \max \{ 4^{2/3}G^2, 
	\frac{4cdG^2}{B_t}+ 6L^2D_1^2 \}$.
	
	Using \cref{eq:h_t,eq:inner_product,eq:bound_on_norm} and taking 
	expectation, we obtain
	\begin{equation}
		\expect[h_{t+1}]\le (1-\frac{1}{T})\expect[h_t]+\frac{1}{T}\left( 
		\frac{\beta_t}{2}\cdot \frac{Q}{(t+4)^{2/3}} +\frac{D_1^2}{2\beta_t}
		\right) + \frac{L}{2 T^2}D_1^2
		\le (1-\frac{1}{T})\expect[h_t]+
		\frac{D_1Q^{1/2}}{T(t+4)^{1/3}}
		+ \frac{L}{2T^2}D_1^2,
	\end{equation}
	where we set $\beta_t = \frac{D_1(t+4)^{1/3}}{Q^{1/2}}$. Using the above 
	inequality recursively, we have
	\begin{align}
	\expect[h_{T+1}]\le & (1-\frac{1}{T})^T (\tF(x_\delta^*)-\tF(\delta\one)) 
	+\sum_{t=1}^T \frac{D_1Q^{1/2}}{T(t+4)^{1/3}} +  \frac{L}{2 T}D_1^2\\
	\le & e^{-1}(\tF(x_\delta^*)-\tF(\delta\one)) + 
	\frac{D_1Q^{1/2}}{T}\int_0^T \frac{\dif x}{(x+4)^{1/3}} + \frac{L}{2 
		T}D_1^2\\
	\leq & e^{-1}(\tF(x_\delta^*)-\tF(\delta\one)) + \frac{D_1Q^{1/2}}{T} 
	\frac{3}{2}(T+4)^{2/3} + \frac{L}{2 T}D_1^2 \\
	\leq & e^{-1}(\tF(x_\delta^*)-\tF(\delta\one))
	+ \frac{D_1Q^{1/2}}{T} \frac{3}{2} (2T)^{2/3} + \frac{L}{2 
		T}D_1^2 \\
	\le & e^{-1}(\tF(x_\delta^*)-\tF(\delta\one)) + 
	\frac{3D_1Q^{1/2}}{T^{1/3}}+ \frac{LD_1^2}{2 T}.
	\end{align}
	By re-arranging the terms, we conclude
	\begin{align}
	(1-\frac{1}{e})\tF(x_\delta^*)-\expect[\tF(z_{T+1})] \le & 
	-e^{-1}\tF(\delta\one)+\frac{3D_1Q^{1/2}}{T^{1/3}}+ \frac{LD_1^2}{2 
		T}\\
	\le & \frac{3D_1Q^{1/2}}{T^{1/3}}+ \frac{LD_1^2}{2 T},
	\end{align}
	where the second inequality holds since 
	the image of $F$ is in $\mathbb{R}_+$.
	
	By \cref{lem:smooth_approx}, we have $\tF(z_{T+1})\le F(z_{T+1})+\delta G$ 
	and 
	\begin{align}
	\tF(x_{\delta}^*)\ge \tF(x^*)-\delta G\sqrt{d}\ge  F(x^*)-\delta 
	G(\sqrt{d}+1).
	\end{align}
	Therefore,
	\begin{align}
	(1-\frac{1}{e})F(x^*)-\expect[F(z_{T+1})] \le  
	\frac{3D_1Q^{1/2}}{T^{1/3}}+ \frac{LD_1^2}{2 T} 
	+ \delta G(1+(\sqrt{d}+1)(1-\frac{1}{e})).%
	\end{align}
\end{proof}

\section{Proof of \cref{thm:zero_stochatic}}\label{app:theorem_zero_stochastic}
\begin{proof}
	By the unbiasedness of $\hat{F}$ and \cref{lem:variance}, we have
	\begin{align}
	\expect[\frac{d}{2\delta}(\hat{F}(y_{t,i}^+)-\hat{F}(y_{t,i}^-))u_{t,i}|\mathcal{F}_{t-1}]
	=& 
	\expect[\expect[\frac{d}{2\delta}(\hat{F}(y_{t,i}^+)-\hat{F}(y_{t,i}^-))u_{t,i}|\mathcal{F}_{t-1},u_{t,i}]|\mathcal{F}_{t-1}]
	\\
	=& 
	\expect[\frac{d}{2\delta}(F(y_{t,i}^+)-F(y_{t,i}^-))u_{t,i}|\mathcal{F}_{t-1}]
	\\
	=& \nabla \tF(z_t),
	\end{align}
	where $z_t= x_t+\delta \one$, and
	\begin{align}
	&\expect[\|\frac{d}{2\delta}(\hat{F}(y_{t,i}^+)-\hat{F}(y_{t,i}^-))u_{t,i}-\nabla
	\tF(z_t) \|^2|\mathcal{F}_{t-1}] \\
	=& \expect[\expect[\| 
	\frac{d}{2\delta}(F(y_{t,i}^+)-F(y_{t,i}^-))u_{t,i}-\nabla \tF(z_t) \\
	&+\frac{d}{2\delta}(\hat{F}(y_{t,i}^+)-F(y_{t,i}^+))u_{t,i} \\
	&- \frac{d}{2\delta} 
	(\hat{F}(y_{t,i}^-)-F(y_{t,i}^-))u_{t,i}\|^2|\mathcal{F}_{t-1},u_{t,i}]|\mathcal{F}_{t-1}]
	\\
	=& \expect[\expect[\| 
	\frac{d}{2\delta}(F(y_{t,i}^+)-F(y_{t,i}^-))u_{t,i}-\nabla 
	\tF(z_t)\|^2|\mathcal{F}_{t-1},u_{t,i}]|\mathcal{F}_{t-1}] \\
	&+ \expect[\expect[\| 
	\frac{d}{2\delta}(\hat{F}(y_{t,i}^+)-F(y_{t,i}^+))u_{t,i}\|^2|\mathcal{F}_{t-1},u_{t,i}]|\mathcal{F}_{t-1}]
	\\
	&+ \expect[\expect[\| \frac{d}{2\delta} 
	(\hat{F}(y_{t,i}^-)-F(y_{t,i}^-))u_{t,i}\|^2|\mathcal{F}_{t-1},u_{t,i}]|\mathcal{F}_{t-1}]
	\\
	\leq& \expect[\| \frac{d}{2\delta}(F(y_{t,i}^+)-F(y_{t,i}^-))u_{t,i}-\nabla 
	\tF(z_t)\|^2|\mathcal{F}_{t-1}] \\
	&+ \frac{d^2}{4\delta^2} \expect[\expect[| 
	\hat{F}(y_{t,i}^+)-F(y_{t,i}^+)|^2\cdot 
	\|u_{t,i}\|^2|\mathcal{F}_{t-1},u_{t,i}]|\mathcal{F}_{t-1}] \\
	&+ \frac{d^2}{4\delta^2} \expect[\expect[| 
	\hat{F}(y_{t,i}^-)-F(y_{t,i}^-)|^2\cdot 
	\|u_{t,i}\|^2|\mathcal{F}_{t-1},u_{t,i}]|\mathcal{F}_{t-1}] \\
	\leq& cdG^2+\frac{d^2}{4\delta^2}\sigma_0^2+\frac{d^2}{4\delta^2}\sigma_0^2 
	\\
	=& cdG^2+\frac{d^2}{2\delta^2}\sigma_0^2.
	\end{align}
	Then we have
	\begin{align}
	\expect[g_t|\mathcal{F}_{t-1}] 
	=&\expect[\frac{1}{B_t}\sum_{i=1}^{B_t}\frac{d}{2\delta}(
	\hat{F}(y_{t,i}^+)-\hat{F}(y_{t,i}^-))u_{t,i}|\mathcal{F}_{t-1}]\\
	=&\nabla \tF(z_t),    
	\end{align}
	and
	\begin{align}
	\expect[\|g_t - \nabla \tF(z_t) \|^2|\mathcal{F}_{t-1}] =&\frac{1}{B_t^2} 
	\sum_{i=1}^{B_t} \expect[\|\frac{d}{2\delta}(
	\hat{F}(y_{t,i}^+)-\hat{F}(y_{t,i}^-))u_{t,i}-\nabla \tF(z_t) 
	\|^2|\mathcal{F}_{t-1}] \\
	\leq& \frac{cdG^2+\frac{d^2}{2\delta^2}\sigma_0^2}{B_t}.     
	\end{align}
	
	Similar to the proof of \cref{thm:zero}, we have
	\begin{equation}
	\expect[\| \nabla \tF(z_t)-\bar{g}_t\|^2]\le 
	\frac{Q}{(t+4)^{2/3}},%
	\end{equation}
	where $Q= \max 
	\{4^{2/3}G^2,6L^2D_1^2+\frac{4cdG^2+2d^2\sigma_0^2/\delta^2}{B_t}  \} 
	$. Thus we conclude
	\begin{align}
	(1-\frac{1}{e})F(x^*)-\expect[F(z_{T+1})] \le  
	\frac{3D_1Q^{1/2}}{T^{1/3}}+ \frac{LD_1^2}{2 T} 
	+ \delta G(1+(\sqrt{d}+1)(1-\frac{1}{e})).
	\end{align}
\end{proof}

\section{Proof of 
	\cref{lem:discrete_to_continuous}}\label{app:lemma_disrite_to_continuous}
\begin{proof}
	Recall that $F(x)=\expect_{X \sim x}[f(X)] = \sum_{S \subseteq 
		\Omega}f(S)\prod_{i \in S}x_i\prod_{j \notin S}(1-x_j)$, then for any 
	fixed 
	$i \in [d]$, where $d=|\Omega|$, we have%
	\begin{align}
	|\frac{\partial F(x)}{\partial x_i}| =& |\sum_{\substack{S \subseteq \Omega 
			\\ i \in S}}f(S)\prod_{\substack{j 
			\in S \\ j \neq i}} x_j \prod_{\substack{k \notin S \\ k \neq i}} 
	(1-x_k)  - \sum_{\substack{S \subseteq \Omega \\ i \notin 
			S}}f(S)\prod_{\substack{j \in S \\ j \neq i}} x_j 
	\prod_{\substack{k \notin 
			S \\ k \neq i}} (1-x_k)| \\
	\le & M[\sum_{\substack{S \subseteq \Omega \\ i \in S}}\prod_{\substack{j 
			\in S \\ j \neq i}} x_j \prod_{\substack{k \notin S \\ k \neq i}} 
	(1-x_k)+ 
	\sum_{\substack{S \subseteq \Omega \\ i \notin S}}\prod_{\substack{j \in S 
			\\ j \neq i}} x_j \prod_{\substack{k \notin S \\ k \neq i}} 
	(1-x_k)] \\
	=& 2M.
	\end{align}
	So we have
	\begin{equation}
	\| \nabla F(x) \| \le 2M\sqrt{d}.   
	\end{equation}
	Then $F$ is $2M\sqrt{d}$-Lipschitz.
	
	Now we turn to prove that $F$ has Lipschitz continuous gradients. Thanks to 
	the multilinearity, we have
	\begin{equation}
	\frac{\partial F}{\partial x_i} = F(x|x_i=1)-F(x|x_i=0).    
	\end{equation}
	Since 
	\begin{equation}
	F(x|x_i=1)=\sum_{\substack{S \subseteq \Omega \\ i \in 
			S}}f(S)\prod_{\substack{j \in S \\ j \neq i}} x_j 
	\prod_{\substack{k \notin 
			S \\ k \neq i}} (1-x_k),    
	\end{equation}
	we have
	\begin{equation}
	\frac{\partial F(x|x_i=1)}{\partial x_i} =0,    
	\end{equation}
	and for any fixed $j \neq i$,
	\begin{align}
	|\frac{\partial F(x|x_i=1)}{\partial x_j}| =& |\sum_{\substack{S \subseteq 
			\Omega \\ i,j \in S}}f(S)\prod_{\substack{l 
			\in S \\ l \notin \{i,j\}}} x_l \prod_{\substack{k \notin S \\ k 
			\notin 
			\{i,j \}}} (1-x_k)  -\sum_{\substack{S \subseteq \Omega \\ i \in S, 
			j \notin 
			S}}f(S)\prod_{\substack{l \in S \\ l \notin \{i,j\}}} x_l 
	\prod_{\substack{k \notin S \\ k \notin \{i,j \}}} (1-x_k)| \\
	\le & M [\sum_{\substack{S \subseteq \Omega \\ i,j \in 
			S}}\prod_{\substack{l \in S \\ l \notin \{i,j\}}} x_l 
	\prod_{\substack{k 
			\notin S \\ k \notin \{i,j \}}} (1-x_k)  + \sum_{\substack{S 
			\subseteq \Omega \\ i \in S, j \notin 
			S}}\prod_{\substack{l \in S \\ l \notin \{i,j\}}} x_l 
	\prod_{\substack{k 
			\notin S \\ k \notin \{i,j \}}} (1-x_k)] \\
	=& 2M.
	\end{align}
	Similarly, we have $\frac{\partial F(x|x_i=0)}{\partial x_i}=0$, and 
	$|\frac{\partial F(x|x_i=0)}{\partial x_j}|\le 2M$ for $j \neq i$. So we 
	conclude that
	\begin{equation}
	|\frac{\partial^2 F}{\partial x_j \partial x_i}| \le 
	\begin{cases}
	0,& \quad \text{if } j=i, \\
	4M,& \quad \text{if } j \neq i.
	\end{cases}
	\end{equation}
	Then $\|\nabla \frac{\partial F}{\partial x_i}\| \le 4M\sqrt{d-1} $, 
	\emph{i.e.}, $\frac{\partial F}{\partial x_i}$ is $4M\sqrt{d-1}$-Lipschitz.

	Then we deduce that
	\begin{align}
	\| \nabla F(z_1) - \nabla F(z_2) \| =& \left[\sum_{i=1}^{d} \left( 
	\frac{\partial F(z_1)}{\partial x_i} - 
	\frac{\partial F(z_2)}{\partial x_i}\right)^2\right]^{1/2} \\
	\le & \left[\sum_{i=1}^{d} (4M\sqrt{d-1})^2 \|z_1-z_2 \|^2\right]^{1/2} \\
	=& \sqrt{\sum_{i=1}^{d} (4M\sqrt{d-1})^2} \cdot \|z_1-z_2 \| \\
	=& 4M\sqrt{d(d-1)} \|z_1-z_2 \|.
	\end{align}
	
	So $F$ is $4M\sqrt{d(d-1)}$-smooth.
\end{proof}

\section{Proof of \cref{thm:zero_discrete}}
\label{app:theorem_discrete}
\begin{proof}
	Recall that we define $z_t= x_t+\delta \one$. Then we have
	\begin{align}
	\expect[\|g_t - \nabla \tF(z_t) \|^2|\mathcal{F}_{t-1}] 
	=& \frac{1}{B_t^2} \sum_{i=1}^{B_t} 
	\expect[\|\frac{d}{2\delta}(\bar{f}_{t,i}^+-\bar{f}_{t,i}^-)u_{t,i}-\nabla 
	\tF(z_t) \|^2|\mathcal{F}_{t-1}] \\
	=&\frac{1}{B_t^2} \sum_{i=1}^{B_t} 
	\expect[\|[\frac{d}{2\delta}(F(y_{t,i}^+)-F(y_{t,i}^-))u_{t,i}-\nabla 
	\tF(z_t)] \\
	&\quad + \frac{d}{2\delta}[\bar{f}_{t,i}^+ - F(y_{t,i}^+)]u_{t,i} - 
	\frac{d}{2\delta} [\bar{f}_{t,i}^- - F(y_{t,i}^-)]u_{t,i} 
	\|^2|\mathcal{F}_{t-1}] \\
	=& \frac{1}{B_t^2} \sum_{i=1}^{B_t} 
	\expect[\|[\frac{d}{2\delta}(F(y_{t,i}^+)-F(y_{t,i}^-))u_{t,i}-\nabla 
	\tF(z_t)]\|^2 | \mathcal{F}_{t-1}] \\
	& \quad + \frac{1}{B_t^2} \sum_{i=1}^{B_t} 
	\expect[|\frac{d}{2\delta}[\bar{f}_{t,i}^+ - F(y_{t,i}^+)] |^2 | 
	\mathcal{F}_{t-1}] \\
	& \quad + \frac{1}{B_t^2} \sum_{i=1}^{B_t} 
	\expect[|\frac{d}{2\delta}[\bar{f}_{t,i}^- - F(y_{t,i}^-)] |^2 | 
	\mathcal{F}_{t-1}],
	\end{align}
	where we used the independence of $\bar{f}^{\pm}_{t,i}$ and the facts that 
	$\expect[\bar{f}^{\pm}_{t,i}] = F(y^\pm_{t,i}), 
	\expect[\frac{d}{2\delta}(F(y_{t,i}^+)-F(y_{t,i}^-))u_{t,i}]=\nabla\tF(z_t)$.

	Then same to \cref{eq:variance_upper} and by 
	\cref{lem:discrete_to_continuous}, the first item is no more than 
	$\frac{4cd^2M^2}{B_t}$. To upper bound the last two items, we have for 
	every $i \in [B_t]$,
	\begin{equation}
	\begin{split}
	\expect[|\frac{d}{2\delta}[\bar{f}_{t,i}^+ - F(y_{t,i}^+)] |^2 | 
	\mathcal{F}_{t-1}] &= \frac{d^2}{4\delta^2} \expect[ 
	[\sum_{j=1}^l 
	[f(Y_{t,i,j}^+) - F(y_{t,i}^+)]/l]^2 |\mathcal{F}_{t-1}] \\
	&\leq \frac{d^2}{4\delta^2}\cdot l \cdot \frac{M^2}{l^2} \\
	&=\frac{d^2M^2}{4l\delta^2}.
	\end{split}  
	\end{equation}

	Similarly, we have 
	\begin{equation}
	\expect[|\frac{d}{2\delta}[\bar{f}_{t,i}^- - F(y_{t,i}^-)] |^2 | 
	\mathcal{F}_{t-1}] \leq \frac{d^2M^2}{4l\delta^2}.
	\end{equation}

As a result, we have 
\begin{equation}
\begin{split}
\expect[\|g_t - \nabla \tF(z_t) \|^2|\mathcal{F}_{t-1}] 
&\le \frac{4cd^2M^2}{B_t} + \frac{1}{B_t^2}\cdot B_t \cdot 
\frac{d^2M^2}{4l\delta^2} + \frac{1}{B_t^2}\cdot B_t \cdot 
\frac{d^2M^2}{4l\delta^2} \\
&= \frac{4cd^2M^2}{B_t} + \frac{d^2M^2}{2B_tl\delta^2}.
\end{split}
\end{equation}

	Then same to the proof for \cref{thm:zero}, we have
	\begin{align}
	(1-\frac{1}{e})F(x^*)-\expect[F(z_{T+1})] \le 
	\frac{3D_1Q^{1/2}}{T^{1/3}}+ \frac{2M\sqrt{d(d-1)}D_1^2}{T} 
	+ 2M\delta 
	\sqrt{d}(1+(\sqrt{d}+1)(1-\frac{1}{e})).\label{eq:discrete_ineq2}
	\end{align}
	where $D_1\triangleq \diam(\constraint')$, $Q= \max \{ 4^{5/3}dM^2, 
	\frac{2d^2M^2(8c+\frac{1}{l\delta^2})}{B_t}+ 96d(d-1)M^2D_1^2 \} $, $x^*$ 
	is the global 
	maximizer of $F$ on $\constraint$.
	
	Note that since the rounding scheme is lossless, we have
	\begin{equation}
	(1-\frac{1}{e})f(X^*) - \expect[f(X_{T+1})] \le  (1-\frac{1}{e})F(x^*) - 
	\expect[F(z_{T+1})].   \label{eq:discrete_ineq1} 
	\end{equation}
	
	Combine \cref{eq:discrete_ineq1,eq:discrete_ineq2}, we have
	\begin{align}
	(1-\frac{1}{e})f(X^*) - \expect[f(X_{T+1})] \le  
	\frac{3D_1Q^{1/2}}{T^{1/3}}+ \frac{2M\sqrt{d(d-1)}D_1^2}{T} 
	+ 2M\delta \sqrt{d}(1+(\sqrt{d}+1)(1-\frac{1}{e})).%
	\end{align}
\end{proof}

\end{document}